\documentclass[10pt,twocolumn,letterpaper]{article}

\usepackage{cvpr}
\usepackage{times}
\usepackage{epsfig}
\usepackage{graphicx}
\usepackage{amsmath}
\usepackage{amssymb}
\usepackage{subcaption} 
\usepackage{multirow}
\usepackage{tabularx}
\usepackage{setspace}
\usepackage[table]{xcolor} 
\usepackage[font={small,it}]{caption} 
\usepackage{url} 
\usepackage{cuted}
\usepackage{capt-of}
\newcommand{\bm}[1]{{\mathbf{#1}}}
\definecolor{greenff}{rgb}{0.0, 0.5, 0.0}
\definecolor{LightCyan}{rgb}{0.88,1,1}

\graphicspath{ {img/} } 
  
\cvprfinalcopy % *** Uncomment this line for the final submission

\ifcvprfinal\fi

\begin{document}
%%%%%%%%% TITLE
\title{Zero-Shot Anticipation for Instructional Activities}

\author{Fadime Sener\\
University of Bonn, Germany\\ 
{\tt\small sener@cs.uni-bonn.de}
\and
Angela Yao\\
National University of Singapore\\ 
{\tt\small ayao@comp.nus.edu.sg}
}

\maketitle

\begin{strip}
\centering 
\vspace{-1.4cm}
 \includegraphics[width=\textwidth]{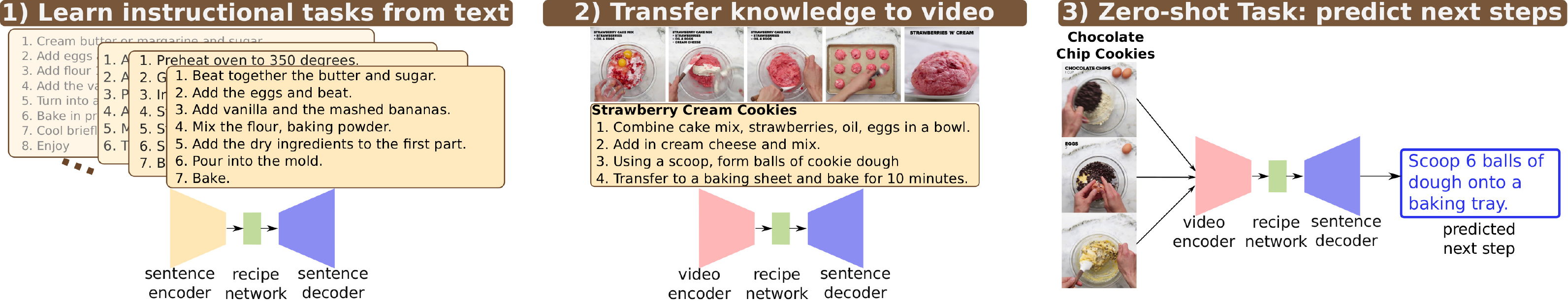}
\vspace{-0.6cm}
\captionof{figure}{We learn procedural knowledge from large text corpora and transfer it to the visual domain to anticipate the future. Our system is composed of four RNNs: a sentence encoder and decoder, a video encoder and a recipe network.
\label{fig:teaser}}
\vspace{-0.27cm}
\end{strip}

\begin{abstract}
\vspace{-0.3cm}
How can we teach a robot to predict what will happen next for an activity it has never seen before? We address this problem of zero-shot anticipation by presenting a hierarchical model that generalizes instructional knowledge from large-scale text-corpora and transfers the knowledge to the visual domain. Given a portion of an instructional video, our model predicts coherent and plausible actions multiple steps into the future, all in rich natural language. To demonstrate the anticipation capabilities of our model, we introduce the \emph{Tasty Videos dataset}, a collection of 2511 recipes for zero-shot learning, recognition and anticipation. 
\end{abstract}

\vspace{-0.9cm}
\section{Introduction} 
Imagine a not-so-distant future, where your kitchen is serviced by a robot chef\footnote{Robots cooking specific recipes~\cite{moley,beetz2011robotic,tenorth2013representation} already exist!}. How should we teach robots to cook? By reading all the recipes on the web? By watching all the cooking videos on YouTube? The ability to learn and generalize from a set of instructions, be it in text, image, or video form, is a highly challenging and open problem faced by those working in machine learning and robotics.

In this work, we limit our scope of training the next robo-chef to predicting subsequent steps as it watches a human cook a never-before-seen dish. We frame our problem as one of future action prediction in a zero- and/or few-shot learning scenario. This best reflects the situation under which service robots will be introduced~{\cite{Chelsea04905,sunderhauf2018limits}}. The robot is pre-trained extensively, but not necessarily with knowledge matching exactly the deployment environment, thereby forcing it to generalize from prior knowledge. At the same time, it is important for the robot to anticipate what will happen in the future, to ensure a safe and smooth collaborative experience with the human~\cite{Koppula15pami,wu2016watch}.

Instructional data and in particular cooking recipes can be readily found on the web~\cite{Instructables,Wikihow}. The richest forms are multimodal, \eg images plus text, or videos with narrations. Such data fits well into our scenario in which the service robot visually recognizes the current context and makes future predictions. However, learning complex, multi-step activities requires significant amounts of data, and despite their online abundance, it is still difficult to find sufficient examples in multi-modal form. Furthermore, learning the visual appearance of specific steps would require temporally aligned data, which is less common and/or expensive to obtain. Our strategy is therefore to separate the procedural learning from the visual appearance learning. Procedural knowledge is learned from text, which is readily available in large corpora on the scale of millions~\cite{salvador2017learning}. This knowledge is then transferred to video, so that the learning of visual appearances can then be simplified to only a grounding model done via aligned video and text (Fig.~\ref{fig:teaser}). More specifically, we encode text and/or video into context vectors. The context is fed to a recipe network, which models the sequential structure of the recipe and makes following step predictions in vector form which are then decoded back into sentences. 

Our work is highly novel in two key regards. First and foremost, we are working with zero-shot action anticipation under a semi-supervised setting, as we target prediction for never-before-seen dishes. We achieve this by generalizing cooking knowledge from large-scale text corpora and then transferring the knowledge to the visual domain. This relieves us of the burden and impracticality of providing annotations for a domain in which there are virtually unlimited number of categories (dishes) and sub-categories (instructional steps). We are the first to tackle such a problem in this form; prior works in complex activity recognition are severely limited in the number of categories and steps~\cite{alayrac2016unsupervised,kuehne2014language,Kuehne12,rohrbach2012database}, while works in action anticipation rely on strong supervision~\cite{abu2018will,lan2014hierarchical,predBerg}. 

Second, we do not work with closed categories derived from word tags; instead we train with and also predict full sentences, \eg \emph{`Cook the chicken wing until both sides are golden brown.'} vs. \emph{`cook chicken'}. This design choice makes our problem significantly more challenging, but also offers several advantages. First of all, it adds richness to the instruction, since natural language conveys much more information than simple text labels~\cite{lin2015generating,zhou2018towards}. It also allows for anticipation of not only actions but also objects and attributes. Finally, as a byproduct, it facilitates data collection, as the number of class-based annotations grows exponentially with the number of actions, objects and attributes and leads to very long-tailed distributions~\cite{Damen2018EPICKITCHENS}.

When transferring knowledge from text recipes to videos, we need to ground the two domains with video with temporally aligned captions. To the best of our knowledge, YoucookII~\cite{zhou2018towards} is currently the only dataset with such labels. However, it lacks diversity in the number of dishes and therefore unique recipe steps. As such, we collect and present our new~\emph{Tasty Videos dataset}, a diverse set of 2511 different cooking recipes\footnote{ Collected from the website \texttt{\url{https://tasty.co/}}} accompanied by a video, ingredient list, and temporally aligned recipe steps. Video footage is taken from a fixed birds-eye view and focuses almost exclusively on the cooking instructions, making it well-suited for understanding the procedural steps. 

We summarize our main contributions as follows: 
\begin{itemize}
 \setlength\itemsep{-0.32em}
 \item We are the first to explore zero-shot action anticipation by generalizing knowledge from large-scale text-corpora and transferring it to the visual domain.
 
 \item We propose a modular hierarchical model for learning multi-step procedures with text and visual context.
 
 \item Our model generalizes cooking knowledge and is able to predict coherent and plausible instructions for multiple steps into the future. The predictions, in rich natural language, score higher in standard NLP metrics than video captioning methods which actually observe the visual data on YouCookII and Tasty Videos.
 
 \item We demonstrate how the proposed approach can be useful for making future step predictions in a zero-shot scenario compared to a supervised setting. 
 
 \item We present a new and highly diverse dataset of 2511 cooking recipes which will be made publicly available and be of interest for those working in anticipation, complex activity recognition and video captioning. 
\end{itemize}
 
\begin{figure*}[htb]
	\centering 
	 	\includegraphics[width=0.99\textwidth]{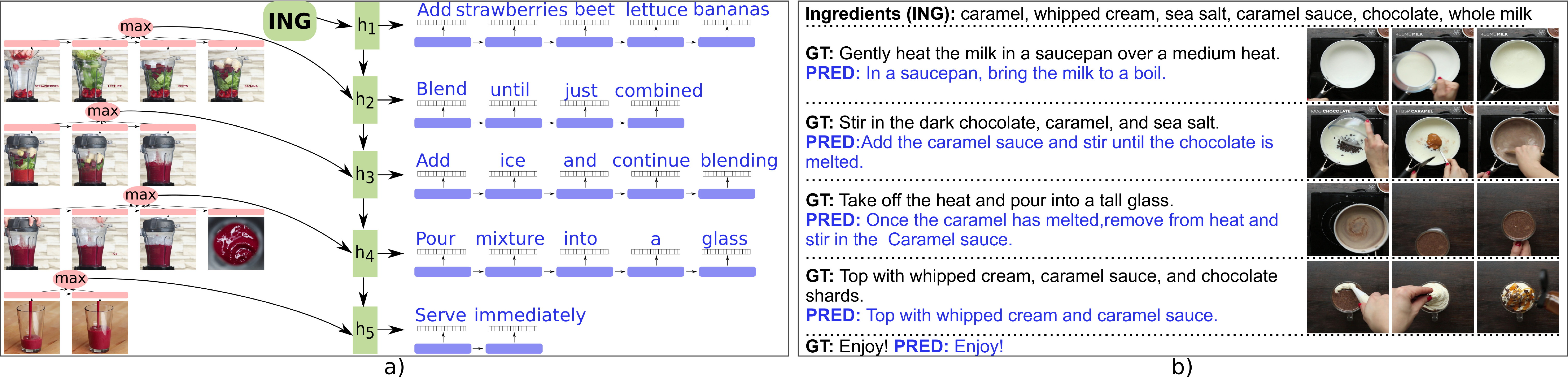}
\vspace{-0.3cm}
\caption{Left: our visual model, composed of video encoder, sentence decoder and recipe RNN. Given the ingredients as initial input and context in visual form, the recipe RNN predicts future steps decoded back into natural language. Right: next step prediction of our visual model. The blue sentences are our model's predictions. Note that our model predicts the next steps before seeing these segments!} 
\label{fig:overview} 
\vspace{-0.5cm}
\end{figure*}

\section{Related Works}
\vspace{-0.1cm}
\textbf{Understanding complex activities} and their sub-activities has been addressed typically as a supervised video segmentation and recognition problem~\cite{kuehne2014language,richard2016temporal,rohrbach2012database}. Newer works are weakly-supervised, using cues from narrations~\cite{malmaud2015s,sener2015unsupervised,alayrac2016unsupervised} or receiving ordered sequences of the actions in videos~\cite{bojanowski2014weakly,huang2016connectionist,richard2017weakly}, or fully unsupervised~\cite{2018_cvpr_sener}. Our work is similar to those using text cues; however, we do not rely on aligned visual-text data for learning the activity models~\cite{alayrac2016unsupervised,sener2015unsupervised} but rather for grounding visual data. 

\textbf{Action prediction} is a new and fast-growing area. Methods for early event recognition~\cite{ryoo2011human, Hoai14ijcv,Xu2015autocompletion} are sometimes (confusingly) also referred to as action prediction, but are incomplete inference methods, since a portion of the action has been observed. Prior work in forecasting activities before making \emph{any} observations have been limited to simple movement primitives ~\cite{Koppula15pami}, or personal interactions ~\cite{lan2014hierarchical,vondrick2016anticipating}. Single predictions are made and the anticipated actions typically occur within a few seconds time frame. Recently,~\cite{abu2018will} predicts multiple actions into the future; our method also predicts multiple steps but unlike~\cite{abu2018will}, we do not require repetitions of activity sequences for training.

\textbf{The cooking domain} is popular in NLP research, since recipes are rich in natural language yet are reasonably limited in scope. Modelling the procedural aspects of text and generating coherent recipes span several decades of work~\cite{DaleRobert88,hammond1986chef,kiddon2016globally,moriMSYHFY14,Morris2012SoupOB}. In multimedia, recipes are involved in tasks such as food recognition~\cite{Herranzfood}, recommender systems~\cite{Mindel2017} and indexing and retrieval~\cite{carvalhoCPSTC18,salvador2017learning}. In computer vision, cooking has been well-explored for complex and fine-grained activity recognition~\cite{Kuehne12,rohrbach2012database,DaXuDoCVPR2013,rohrbach14gcpr,Damen2018EPICKITCHENS,zhou2018towards}, temporal segmentation~\cite{Kuehne12,zhou2018towards} and captioning~\cite{rohrbach2013translating,regneri2013grounding,zhou2018end}. Several cooking and kitchen-related datasets have been presented~\cite{Damen2018EPICKITCHENS,malmaud2015s,rohrbach2012database, Kuehne12,zhou2018towards} and feature a wide variety of labels depending on the task. Two~\cite{malmaud2015s,zhou2018towards} are similar to our new dataset, in that they include recipe texts and accompanying videos. However, YouCookII~\cite{zhou2018towards} has limited diversity in activities with only 89 dishes; \cite{malmaud2015s} is larger in scale, but lacks temporal alignments between texts and videos. 
  
\vspace{-0.1cm}
\section{Modelling Sequential Instructions}\label{sec:model}
\vspace{-0.1cm}
Sequence-to-sequence learning~\cite{sutskever2014sequence} has made it possible to successfully generate continuous text and build dialogue systems~\cite{cho2014learning,vinyals2015neural}. 
Recurrent neural networks (RNNs) are used to learn rich representations of sentences~\cite{hill2016learning,Ba2016LayerN,sThought15} in an unsupervised manner, using the extensive amount of text that exists in books and web corpuses. However, for instructional text such as cooking recipes, such representations tend to do poorly, and suffer from coherence from one time step to the next, since they do not fully capture the underlying sequential nature of the instruction set. As such, we propose a hierarchical model with four components, where the sentences and the steps of the recipe are represented by two dedicated RNNs: the sentence encoder and the recipe RNN respectively. A third RNN decodes predicted recipe steps back into sentence form for human-interpretable results (sentence decoder). These three RNNs are learned jointly as an auto-encoder in an initial training step. A fourth RNN encoding visual evidence (video encoder) is then learned in a subsequent step to replace the sentence encoder to enable interpretation and future prediction from video. An overview is shown in Fig.~\ref{fig:teaser}, while details of the RNNs are given in Sections~\ref{sec:sentence} to~\ref{sec:videoenc}.
 
\vspace{-0.04cm}
\subsection{Sentence Encoder and Decoder}~\label{sec:sentence}
The sentence encoder produces a fixed-length vector representation of each recipe step. We use a bi-directional LSTM and following ~\cite{conneauEMNLP2017} we apply a max pooling over each dimension of the hidden units. More formally, let sentence $s_{j}$ from step $j$ of a recipe (we assume each step is one sentence) be represented by $M$ words, \ie $s_{j} = \{w_{j}^{t}\}_{t=1...M}$ and ${\bf{x}}_{j}^t$ be the word embedding of word $w_{j}^{t}$. For each sentence $j$, at each (word) step $t$, the bi-directional $\text{LSTM}_{\text{se}}$ outputs $\bm{y}_{j}^t$, where
\vspace{-0.09cm}
\begin{equation}\label{eq:sentenc}
 \!\!\!\!\! {\bm{y}_{j}^t} \! = \! \left[\text{LSTM}_{\text{se}}\left(\{\bm{x}_{j}^{1}, ..., \bm{x}_{j}^{t}\}\right)\!, \text{LSTM}_{\text{se}}\left(\{\bm{x}_{j}^{M}, ..., \bm{x}_{j}^{t}\}\right)\right] 
\end{equation}
\noindent which is a concatenation of the hidden states from the forwards and backwards pass of $\text{LSTM}_{\text{se}}$. The overall sentence representation $\bm{r}_{j}$ is determined by a dimension-independent max-pooling over the time steps, \ie
\vspace{-0.09cm}
\begin{equation}\label{eq:maxpool}
( \bm{r}_{j})_d = \max_{t \in \{1,...,M\} } (\bm{y}_{j}^t)_d,
\end{equation}
where $(\cdot)_d,\ d\!\in\!\{1, ..., D\}$ indicates the $d$-th element of the $D$-dimensional bi-directional LSTM outputs $\bm{y}_{j}^t$. The decoder is an LSTM-based neural language model which converts the fixed-length representation of the steps back into sentences. More specifically, given the prediction $\hat{\bm{r}}_{j}$ from the recipe RNN of step $j$, it decodes the sentence $\hat{s}_{j}$
\vspace{-0.09cm}
\begin{equation}
 \hat{s}_{j} = \text{LSTM}_\text{d}(\hat{\bm{r}}_{j}) = \{\hat{w}_{j}^1, ..., \hat{w}_{j}^{\hat{M}}\}.
\end{equation}
 
\subsection{Recipe RNN}~\label{sec:recipeRNN}
We model the sequential ordering of recipe steps with an LSTM which takes as input $\{\bm{r}_{j}\}_{j=1,...,N}$, \ie fixed-length representations of the steps of a recipe with $N$ steps, where $j$ indicates the step index. At each (recipe) step, the hidden state of the recipe RNN $\bm{h}_{j}$ can be considered a fixed-length representation of all recipe steps $\{s_{1}, ..., s_{j}\}$ seen up to step $j$; we directly use this hidden state vector as a prediction of the sentence representation for step $j+1$, \ie
\vspace{-0.2cm}
\begin{equation}
\hat{\bm{r}}_{j+1} = \bm{h}_{j} = \text{LSTM}_{\text{r}}(\{\bm{r}_{0},...,\bm{r}_{j}\}).
\end{equation}

\noindent 
The hidden state of the last step ${\bm{h}}_{N}$ can be considered a representation of the entire recipe. Due to the standard recursion of the hidden states in $\text{LSTM}_\text{r}$, each hidden state vector and therefore each future step prediction is conditioned on the previous steps. This allows to predict recipe steps which are plausible and coherent with respect to previous steps. 

Recipes usually include an ingredient list which is a rich source of information that can also serve as a strong modelling cue~\cite{kiddon2016globally,salvador2017learning}. To incorporate the ingredients, we form an ingredient vector $\bm{I}$ for each recipe in the form of a one-hot encoding over a vocabulary of ingredients. $\bm{I}$ is then transformed with a separate fully connected layer in the recipe RNN to serve as the initial input, \ie $\bm{r}_{0} = f(\bm{I}).$

\subsection{Video Encoder}\label{sec:videoenc}
For inference, we would like the recipe RNN to interpret sentences from text inputs and also visual evidence. Due to the modular nature of our proposed model, we can conveniently replace the sentence encoder with an analogous video encoder. Suppose the $j^{\text{th}}$ video segment $c_{j}$ is composed of $L$ frames, \ie $c_{j} = \{\bm{f}_{j}^t\}_{t=1,...,L}$. Each frame $f_{j}^t$ is represented as a high-level CNN feature vector -- we use the last fully connected layer output of ResNet-50~\cite{he2016deep} before the softmax layer. Similar to the sentence encoding $\bm{r}_{j}$ in Eqs.~\ref{eq:sentenc} and~\ref{eq:maxpool}, we determine the video encoding vector $\bm{v}_{j}$ by applying a dimension-independent max pooling over the time steps of $\bm{z}_{j}^t$, where : 
\begin{equation}
\!\!\!\!\!\! {\bm{z}_{j}^t} \! = \! \left[\text{LSTM}_{\text{ve}}\!\left(\{\bm{f}_{j}^{1}, ..., \bm{f}_{j}^{t}\}\right)\!, \text{LSTM}_{\text{ve}}\!\left(\{\bm{f}_{j}^{M}, ..., \bm{f}_{j}^{t}\}\right)\right]. 
\end{equation}
\noindent The video encoding $\text{LSTM}_{\text{ve}}$ is trained such that $\bm{v}_{j}$ can directly replace $\bm{r}_{j}$, as detailed in the following.

\vspace{-0.2cm}
\subsection{Model Learning and Inference}
The full model is learned in two stages. First, the sentence encoder ($\text{LSTM}_{\text{se}}$), recipe RNN ($\text{LSTM}_{\text{r}}$) and sentence decoder ($\text{LSTM}_{\text{d}}$) are jointly trained end-to-end. Given a recipe of $N$ steps, a loss can be defined as the negative log probability of each reconstructed word:
\vspace{-0.4cm}
\begin{equation}\label{eq:objective}
 L(s_1, ..., s_{N}) = - \sum_{j=1}^{N} \sum_{t=1}^{M_j} \log P( w_{j}^{t} | w_{j}^{t'<t}, \hat{\bm{r}}_j),
 \vspace{-0.4cm}
\end{equation}
where $P( w_{j}^{t} | w_{j}^{t'<t}, \hat{\bm{r}}_j)$ is parameterised by a softmax function at the output layer of the sentence decoder to estimate the distribution over the words $w$ in our vocabulary $V$. The overall objective is then summed over all recipes in the corpus. The loss is computed only when the LSTM is learning to decode a sentence. This first training stage is unsupervised, as the sentence encoder and decoder and the recipe RNN require only text inputs which can easily be scraped from the web without human annotations. In a second step, we train the video encoder ($\text{LSTM}_{\text{ve}}$) while keeping the recipe RNN and sentence decoder fixed. We simply replace the sentence encoder with the video encoder while applying the same loss function as defined in Eq.~\ref{eq:objective}. This step is supervised, as it requires video segments of each step temporally aligned with the corresponding sentences. 

During inference, we provide the ingredient vector $\bm{r}_{0}$ as an initial input to the recipe RNN, which then outputs the predicted vector $\hat{\bm{r}}_{1}$ for the first step (see Fig.~\ref{fig:overview}). We use the sentence decoder and generate the first sentence $\hat{s}_{1}$. Then, we sample a sequence of frames from the video and apply the video encoder to generate $\bm{v}_{1}$ which we again provide as input to the recipe RNN. The output prediction of the recipe RNN, $\hat{\bm{r}}_{2}$, is for the second step of the video. We again use the sentence decoder and generate the next sentence $\hat{s}_{2}$.

Our model is not limited to one step ahead predictions: for further predictions, we can simply apply the predicted output $\hat{\bm{r}}_{j}$ as contextual input ${\bm{r}}_{j}$. During training, instead of always feeding in the ground truth ${\bm{r}}_{j}$, we sometimes (with 0.5 probability after the 5th epoch) use our predictions $\hat{\bm{r}}_{j}$ as the input for the next step predictions that helps us with being robust to feeding in bad predictions~\cite{bengio2015scheduled}. 

\vspace{-0.1cm}
\subsection{Implementation and Training Details}
\vspace{-0.1cm}
We use a vocabulary $V$ of 30171 words provided by Recipe1M~\cite{salvador2017learning}; words are represented by 256-dimensional vectors shared between the sentence encoder and decoder. Our ingredients vocabulary has 3769 ingredients; the one-hot ingredient encodings are mapped into a 1024 dimensional vector $\bm{r}_0$. The RNNs are all single-layer LSTMs implemented in PyTorch; $\text{LSTM}_{\text{se}}$, $\text{LSTM}_{\text{ve}}$, $\text{LSTM}_{\text{d}}$ have 512 hidden units while $\text{LSTM}_{\text{d}}$ has 1024. We train our model using the Adam optimizer~\cite{kingma2014adam} with a batch size of 50 recipes and a learning rate of 0.001; the text-based model is trained for 50 epochs and the visual encoder for 25 epochs.

\begin{figure*}[htb]
\footnotesize
\bgroup
\def\arraystretch{1.25}
\setlength\tabcolsep{0.25em}
\setstretch{0.1}
\begin{tabularx}{0.99\textwidth}{|c|X|X|c|c|c|c|c|}
\hline
& \textbf{ground truth~(GT)} & \textbf{prediction} & \tiny{\textbf{BLEU1}} & \tiny{\textbf{BLEU4}} & \tiny{\textbf{METEOR}} & \tiny{\textbf{HUMAN1}}& \tiny{\textbf{HUMAN2}}\\\hline
\textbf{ING} & bacon, brown sugar, cooking spray, breadsticks & & & & & & \\\hline 
\textbf{step1} & Preheat oven to 325 degrees F ( 165 degrees C ). & Preheat oven to 400 degrees F. & \cellcolor{orange!36}36.0 & \cellcolor{red!0}0.0 & \cellcolor{blue!26} 26.0 & \cellcolor{greenff!50} 1.5 & \cellcolor{greenff!50} 1.5 \\\hline
\textbf{step2} & Line 2 baking sheets with aluminum foil or parchment paper and spray with cooking spray.	& Line a baking sheet with aluminum foil. & \cellcolor{orange!23}23.0 & \cellcolor{red!0}0.0 & \cellcolor{blue!23} 23.0 & \cellcolor{greenff!25}1.0 & \cellcolor{greenff!25}1.0\\\hline
\textbf{step3} & Wrap 1 bacon strip around each breadstick, leaving about 1 inch uncovered on each end. & Place bacon strips in a single layer on the prepared baking sheet.&\cellcolor{orange!13} 13.0 & \cellcolor{red!0}0.0 & \cellcolor{blue!9}9.0 & \cellcolor{greenff!10}0.5 & \cellcolor{greenff!50} 1.5 \\\hline
\textbf{step4} & Place wrapped breadsticks on the prepared baking sheet. & Place rolls on a baking sheet. & \cellcolor{orange!48}48.0 & \cellcolor{red!0}0.0 &\cellcolor{blue!30}30.0 & \cellcolor{greenff!50} 1.5 & \cellcolor{greenff!50} 1.5 \\\hline
\cellcolor{lime!8}\textbf{step5} & \cellcolor{lime!8}Sprinkle brown sugar evenly over breadsticks. &\cellcolor{lime!8} Bake in the preheated oven until breadsticks are golden brown, about 15 minutes. & \cellcolor{orange!15}15.0 & \cellcolor{red!0}0.0 &\cellcolor{blue!13}13.0 & \cellcolor{greenff!0}0.0 & \cellcolor{greenff!50} 1.5\\\hline
\cellcolor{blue!5}\textbf{step6} & \cellcolor{blue!5}Bake in the preheated oven until bacon is crisp and browned, 50 to 60 minutes. & \cellcolor{blue!5}Bake in preheated oven until bacon is crisp and breadsticks are golden brown, about 15 minutes.& \cellcolor{orange!63}63.0 & \cellcolor{red!43}43.0 &\cellcolor{blue!36}36.0 & \cellcolor{greenff!25}1.0 & \cellcolor{greenff!25}1.0\\\hline 
\textbf{step7} & Cool breadsticks on a piece of parchment paper or waxed paper sprayed with cooking spray. & Remove from oven and let cool for 5 minutes.&\cellcolor{orange!6} 6.0 & \cellcolor{red!0}0.0 &\cellcolor{blue!4}4.0 & \cellcolor{greenff!10}0.5 & \cellcolor{greenff!50} 1.5 \\\hline
\end{tabularx}
\egroup
\vspace{-2.5mm}
\caption{Predictions of our text-based method for ``Candied Bacon Sticks'' along with the automated scores and human ratings. For ``HUMAN1'' we ask the raters to directly assess how well the predicted steps match the corresponding ground truth~(GT) sentences, for ``HUMAN2'' we ask to judge if the predicted step is still a plausible future prediction, see Sec.~\ref{sec:human_study}. Our prediction for step 6 matches the GT well while step 5 does not. However, according to ``HUMAN2'' score, our step 5 prediction is still a plausible future action. } 
\label{fig:example_preds}
\vspace{-0.55cm}
\end{figure*}

\vspace{-0.1cm}
\section{Tasty Videos Dataset}\label{sec:tastydataset}
\vspace{-0.1cm}
Our new \emph{Tasty Videos Dataset} has 2511 unique recipes collected from the Buzzfeed website \texttt{\small\url{https://tasty.co}}. Each recipe has an ingredient list, step-wise instructions and a video demonstrating the preparation. The recipes feature breakfast, dinner, desserts, and drinks from 185 categories such as cakes, pies, soups. We define a split ratio of 8:1:1 for training, validation and testing, each containing different recipes. Our test setting is therefore zero-shot, as we make predictions on unseen recipes. We further divide the test set into recipes with similarities in the training set, \eg \emph{``Strawberry Breakfast Muffins''} vs.\ \emph{``Carrot Cake Muffins''} and those without any similarities \eg \emph{``Pigs In A Blanket''}.

The Tasty Videos are captured with a fixed overhead camera and focus entirely on preparation of the dish (see Fig.~\ref{fig:overview}). This viewpoint removes the added challenge of distractors and irrelevant actions and while it may not exactly reflect the visual environments one may find in the home, this simplification allows us to focus the scope of our work on modelling the sequential nature of instructional data, which is already a highly challenging and open research topic. The videos are short (on average 1551 frames / 54 seconds) yet contain a challenging number of steps (9 on average). For each recipe step, we annotate the temporal boundaries in which the step occurs within the video, omitting those without visual correspondences, such as alternative recommendations, non-visualized instructions such as \emph{`Preheat oven.'}\ and stylistic statements such as \emph{`Enjoy!'}.

\vspace{-0.15cm}
\section{Experiments}
\vspace{-0.05cm}
\subsection{Datasets and Evaluation Measures}
\vspace{-0.1cm}
We train and evaluate our method with Recipe1M~\cite{salvador2017learning}, YoucookII~\cite{zhou2018towards} and our Tasty Videos. Recipe1M features approximately one million text recipes with a dish name, list of ingredients, and sequence of instructions. YoucookII is a collection of 2000 cooking videos from YouTube from 89 dishes annotated with the temporal boundaries of each step. We use the ingredients and instructions from the Recipe1M training split to learn our sentence encoder, decoder and recipe RNN. To learn the video encoder, we use the aligned instructions and video data from the training split of either YouCookII or Tasty Videos. We evaluate our model's prediction capabilities with text inputs from Recipe1M and video and text inputs from YoucookII and Tasty Videos. 

Our predictions are in sentence form; evaluating the quality of generated sentences is known to be difficult in natural language processing~\cite{vedantam2015cider,lopez2008statistical}. We apply a variety of evaluation measures in order to offer a broad assessment. First, we target the matching of ingredients and verbs, since they indicate the next active objects and actions and are analogous to the assessments made in action anticipation~\cite{Damen2018EPICKITCHENS}. Second, we evaluate with sentence matching scores BLEU~\cite{papineni2002bleu} and METEOR~\cite{banerjee2005meteor} which are also used for video captioning methods~\cite{regneri2013grounding,rohrbach2013translating,zhou2018end}. Note that automated scores are best at indicating precise word matches to ground truth~(GT) and often do not match sentences a human would consider equivalent. We therefore conduct a user study and ask people to assess how well the predicted step matches the GT in meaning; if it does not match, we ask if the prediction would be plausible for future steps. This gives flexibility in case predictions do not follow the exact aligned order of the ground truth, \eg due to missing steps not predicted, or steps which are slightly out of order (see Fig.~\ref{fig:example_preds}) 
 
\begin{figure*}[ht]
\centering 
 \includegraphics[width=0.99\textwidth]{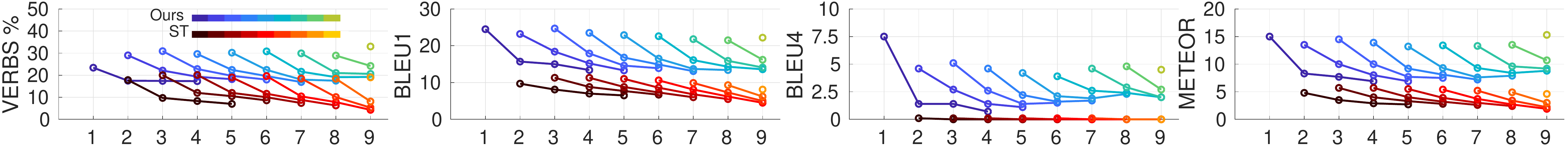}
\vspace{-2.0mm}
\caption{The recall of verbs and sentence scores computed between the predicted and the ground truth sentences for our model (Ours) and the skip-thoughts (ST) model. The x-axes in the plots indicate the step number being predicted in the recipe; each curve begins on the first (relative) prediction, \ie the $(j+1){\text{th}}$ step after having received steps $1$ to $j$ as input.} 
\label{fig:sents_ours_st_9} 
\vspace{-0.35cm}
\end{figure*} 

\begin{figure}[ht]
\centering 
 \includegraphics[width=0.99\columnwidth]{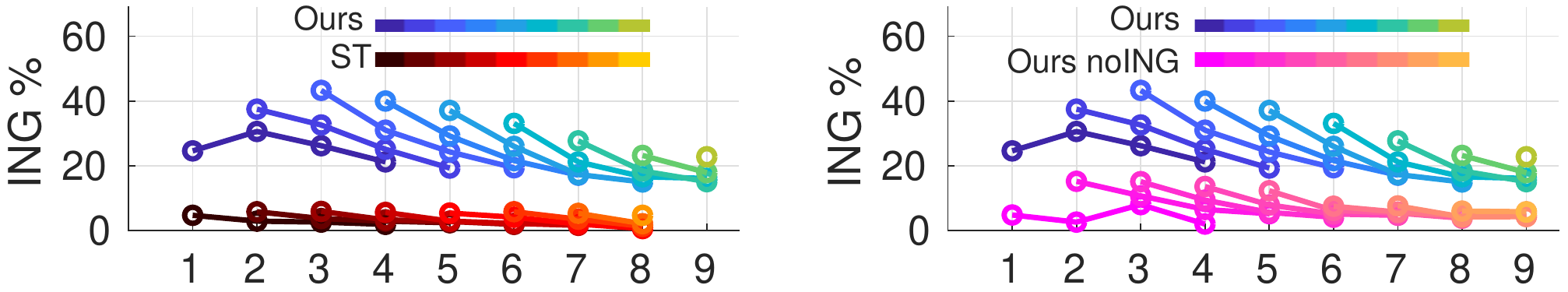} 
\vspace{-2.0mm}
\caption{The recall of ingredients predicted by our model (Ours), by our model trained without the ingredients (Ours noING) and the by skip-thoughts model (ST). The x-axes indicate the step number being predicted in the recipe. 
} 
\label{fig:inredients_ours_st} 
\vspace{-0.5cm}
\end{figure} 

\vspace{-0.2cm}
\subsection{Learning of Procedural Knowledge}
\vspace{-0.1cm}
We first verify the learning of procedural knowledge with a text-only model, evaluating on Recipe1M's test set of 51K recipes. For a recipe of $N$ steps, we predict steps $j\!+\!1$ to $N$, conditioning on steps 1 to $j$ as input context. $N$ varies from recipe to recipe so we separately tally recipes with $N\!\!=\!\!9$ (4300 recipes; 9 is also the average number of steps in the test set) which we report here. Results over the entire test set follow similar trends and are shown in the Supplementary. 
 
For comparison, we look at the generations from a skip-thought (ST) model~\cite{sThought15}. Skip-thought models are trained to decode temporally adjacent sentences from a current encoding, \ie given step $j$ to the encoder, the decoder predicts step $j\!+\!1$, and have been shown to be successful in generating continuous text~\cite{cho2014learning,vinyals2015neural,kiddon2016globally}. We train the ST model on the training set of the Recipe1M dataset. Because the ST model generations are not trained to accept an ingredient list as a $0^{\text{th}}$ or initialization step, we make ST predictions only from the second step on-wards.

\textbf{Key Ingredients:} 
We first compare the recall of ingredients in our predictions to an ST model and a variation of our model trained without ingredients. Rather than directly cross-referencing the ingredient list, we limit the evaluation to ingredients mentioned explicitly in the recipe steps. This is necessary to avoid ambiguities that may arise from specific instructions such as \emph{`add chicken, onion, and bell pepper'} versus the more vague \emph{`add remaining ingredients'}. Furthermore, the ingredient lists in Recipe1M are often automatically generated and may be incomplete. Fig.~\ref{fig:inredients_ours_st} shows that our model's predictions successfully incorporate relevant ingredients with recall rates as high as 43.3\% with the predicted (relative) next step. The overall recall decreases with the (absolute) latter steps, likely due to increased difficulty once the overall number of ingredient occurrences decreases, which tends to happen in later steps. 

Compared to the ST model, our predictions' ingredient recall is higher regardless of whether or not ingredients are provided as an initial input. Without ingredient input, the overall recall is lower but after the initial step, our model's recall increases sharply, \ie once it receives some context. We attribute this to the strength of our model to generalize across related recipes, so that it is able to predict relevant co-occurring ingredients. Our predictions include common ingredients such as \emph{salt, butter, eggs} and \emph{water} and also recipe-specific ones such as \emph{couscous, zucchini}, or \emph{chocolate chips}. While the ST model predicts some common ingredients, it fails to predict recipe-specific ingredients.

\textbf{Key verbs} indicate the main action for a step and are also cues for future steps both immediate (\eg after \emph{`adding'} ingredients into a bowl, a common next step may be to \emph{`mix'}) and long-term (\eg after \emph{`preheating'} the oven, one expects to \emph{`bake'}). We tag the verbs in the training recipes with the Natural Language Toolkit~\cite{nltk} and select the 250 most frequent for evaluation. Similar to ingredients, we check for recall of these verbs only if they appear in the ground truth steps. In the ground truth, there are between 1.55 and 1.85 verbs per step, \ie steps often include multiple verbs such as \emph{``add and mix''}. Fig.~\ref{fig:sents_ours_st_9} shows that our model recalls up to 30.9\% of the verbs with the predicted next step. Our performance is worst in the first (absolute) steps, due to ambiguities when given only the ingredients without any further knowledge of the recipe. After the first steps, our performance quickly increases and stays consistent across the remaining steps. In comparison, the ST model's best recall is only 20.1\% for the next step prediction.

\textbf{Sentences:} 
Our model is able to predict coherent and plausible instructional sentences as shown in Fig.~\ref{fig:example_preds}; more predictions can be found in the Supplementary Materials. We evaluate the entire predicted sentences with BLEU1, BLEU4 and METEOR scores (see Fig~\ref{fig:sents_ours_st_9}). For our model, the BLEU1 scores are consistently high, at around 25.0 for the next (relative) step predictions, with a slight decrease towards the end of the recipes. Predictions further than the next step have lower scores, though they stay above 15.0. The BLEU4 scores are highest in the very first step and range between 1.0 and 5.0 over the remaining steps. The high scores at the early steps are because many recipes start with common instructions such as \emph{`Preheat oven to X degrees'} or \emph{`In a large skillet, heat the oil'}. Similarly, we also do well towards the end of recipes, where instructions for serving and garnishing are common, \eg, \emph{`Season with salt and pepper.'}. Trends for the METEOR score are similar.

Our method outperforms the ST model across the board. In fact, predictions up to four steps into the future surpass the ST predictions only one step ahead. This can be attributed to the dedicated long-term modelling of the recipe RNN that allows us to incorporate the context from all sentence inputs up to the present. In contrast, ST are Markovian in nature and can only take the current step into account. 

In cooking recipes, one does not only find strict instructional steps, but also suggestions based on experience. An interesting outcome of our model is that it also makes such recommendations. For example, for the ground truth \emph{`If it's too loose place it in the freezer for a little while to freeze.'}, our model predicts \emph{`If you freeze, it will be easier to eat'}.

\vspace{-0.1cm}
\subsection{Video Predictions}
\vspace{-0.1cm}
We evaluate our model for making predictions on video inputs on YouCookII's validation set and Tasty Videos' test set. We test two video segmentation settings for inference: one according to ground truth (Ours Visual (GT)) and one based on fixed windows (Ours Visual). In both settings, we sample every fifth frame in these segments and feed their visual features into the recipe RNN as context vectors. Compared to using ground truth segments, the fixed window segments do not have a significant decrease in performance (5\%-18\%,see Tables~\ref{fig:tasty_res} and ~\ref{fig:YoucookII_res_supervised} for Tasty and YouCookII respectively). Overall, our method is relatively robust to the window size (see Supplementary) and we report here results for a window of 70 frames for YouCookII and 170 for Tasty. 
 
\begin{figure*}[ht]
\centering 
	\includegraphics[width=\textwidth]{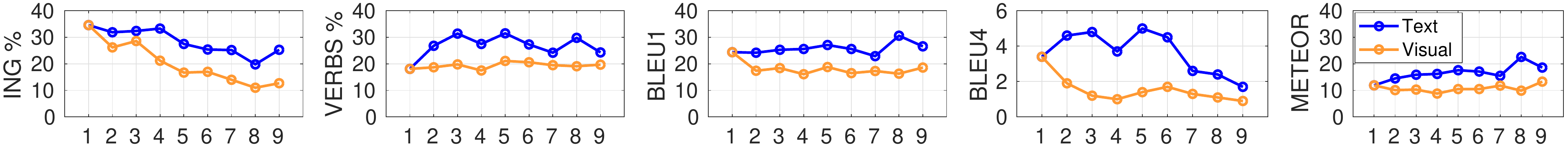}
\vspace{-0.7cm}
\caption{Our results on the Tasty dataset for next step predictions only for our visual and textual model for the recall of predicted ingredients and verbs, and sentence scores. Compared to text, our visual model has a lower performance, but follow similar trends.} 
\label{fig:tasty_all_res} 
\vspace{-0.44cm}
\end{figure*}

Through the video encoder, our model can interpret visual evidence and make plausible predictions of next steps (see examples in Fig.~\ref{fig:overview}(b) and \ref{fig:video_correction}, more results in Supplementary). Given that the model is first trained on text and then transferred to video, the drop in performance from text to video is as expected, though video results still follow similar trends (see Fig.~\ref{fig:tasty_all_res}, compare to ``Ours Text'' in Tables~\ref{fig:tasty_res} and~\ref{fig:YoucookII_res_supervised} for Tasty Videos and YouCookII respectively). 
 
We further investigate the influence of the ingredients on the performance of our method. The performance decrease is mainly noticeable in the ingredient scores and the BLUE4 scores. When ingredients are not provided, our method fails to make plausible predictions in the early stages. After the initial steps, our method receives enough context and the scores increase, see Supplementary.

In some instructional scenarios, there may be semi-aligned text that accompanies video, \eg narrations. We test such a setting by training the sentence and video encoder, as well as sentence decoder and recipe RNN jointly for making future step predictions. We concatenate the sentence and video context vectors and then pass them through a linear layer before feeding them as input to the Recipe RNN, and observe that the results are better than our video alone results but not better than our text alone results (see ``Ours joint video-text'' in Table~\ref{fig:tasty_res}). Even with joint training, it is still difficult to make improvements, which we attribute to the diversity in our videos and variations in the text descriptions for similar visual inputs. On the other hand, when there is accompanying text, our model can be adapted easily and improves the prediction performance. 

\begin{figure}[ht]
\vspace{-0.2cm}
\centering 
\includegraphics[width=0.97\columnwidth]{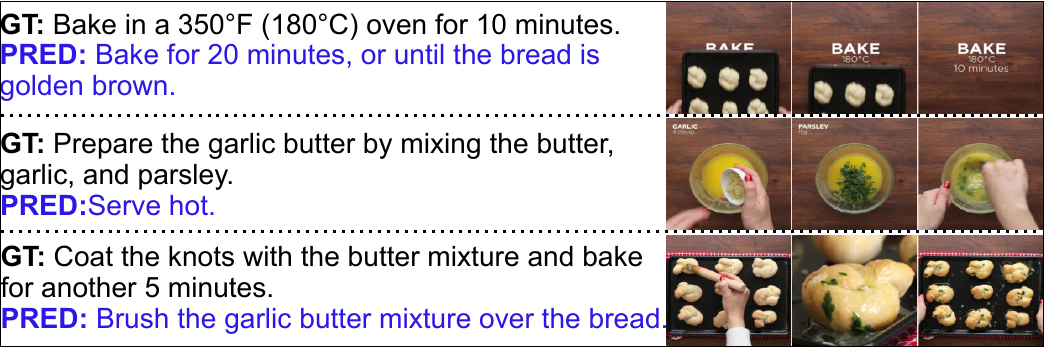} 
\vspace{-0.3cm}
\caption{Next step prediction of our visual model: blue sentences are our model's predictions. After baking, our model predicts that the dish should be served, but after visually seeing the butter parsley mixture, it predicts that the knots should be brushed.
}
\label{fig:video_correction} 
\vspace{-0.35cm}
\end{figure} 

\begin{table}[ht]
\footnotesize
\bgroup
\def\arraystretch{1.25}
\setlength\tabcolsep{0.748em}
\setstretch{0.4}
\begin{tabularx}{0.99\columnwidth}{|c|r|r|r|r|r|}
\hline 
\scriptsize{Method} & \tiny{ING} & \tiny{VERBS} & \tiny{BLEU1} & \tiny{BLEU4} & \tiny{METEOR} \\\hline 
S2VT~\cite{venugopalan2015} (GT) &\cellcolor[HTML]{FFD0E9}7.59&\cellcolor[HTML]{BBEEBB}19.18 &\cellcolor[HTML]{FFD7AE}18.03 &\cellcolor[HTML]{FFC6C6}1.10 &\cellcolor[HTML]{C1C1F9}9.12 \\ 
S2VT~\cite{venugopalan2015}, next (GT)&\cellcolor[HTML]{FFEDF7}1.54&\cellcolor[HTML]{EEFBEE}10.66 &\cellcolor[HTML]{FFF5EC}9.14 &\cellcolor[HTML]{FFEBEB}0.26 &\cellcolor[HTML]{ECECFD}5.59 \\ 
End-to-end~\cite{zhou2018end} & - & - & - &\cellcolor[HTML]{FFEBEB}0.54 &\cellcolor[HTML]{ECECFD}5.48 \\\hline
Ours Visual (GT) &\cellcolor[HTML]{FF7AC2}20.40 &\cellcolor[HTML]{BBEEBB}19.18 &\cellcolor[HTML]{FFD1A3}19.05 &\cellcolor[HTML]{FFB2B2}1.48 &\cellcolor[HTML]{9C9CF5}11.78 \\ 
Ours Visual &\cellcolor[HTML]{FF95CE}16.66 &\cellcolor[HTML]{CCF2CC}17.08 &\cellcolor[HTML]{FFD9B3}17.59 &\cellcolor[HTML]{FFBFBF}1.23 &\cellcolor[HTML]{A7A7F6}11.00 \\\hline
Ours Text (100\%) &\cellcolor[HTML]{FF4FAE}26.09 &\cellcolor[HTML]{65DA65}27.19 &\cellcolor[HTML]{FF9F40}26.78 &\cellcolor[HTML]{FF4040}3.30 &\cellcolor[HTML]{4040EC}17.97 \\ 
Ours Text (50\%) &\cellcolor[HTML]{FF66B9}23.01 &\cellcolor[HTML]{81E081}24.90 &\cellcolor[HTML]{FFAC58}25.05 &\cellcolor[HTML]{FF7979}2.42 &\cellcolor[HTML]{4F4FED}16.98 \\ 
Ours Text (25\%) &\cellcolor[HTML]{FF81C5}19.43 &\cellcolor[HTML]{8DE38D}23.83 &\cellcolor[HTML]{FFB66D}23.54 &\cellcolor[HTML]{FF9191}2.03 &\cellcolor[HTML]{5D5DEF}16.05 \\ 
Ours Text (0\%) &\cellcolor[HTML]{FFDAEE}5.80 &\cellcolor[HTML]{F0FBF0}9.42 &\cellcolor[HTML]{FFF3E8}10.58 &\cellcolor[HTML]{FFECEC}0.24 &\cellcolor[HTML]{DFDFFC}6.80 \\\hline
Ours Text noING &\cellcolor[HTML]{FFC7E5}9.04 &\cellcolor[HTML]{A0E8A0}22.00 &\cellcolor[HTML]{FFCB97}20.11 &\cellcolor[HTML]{FFD0D0}0.92 &\cellcolor[HTML]{8989F3}13.07 \\\hline
Ours joint video-text &\cellcolor[HTML]{FF6CBB}22.27 &\cellcolor[HTML]{92E492}23.35 &\cellcolor[HTML]{FFC184}21.75 &\cellcolor[HTML]{FF7F7F}2.33 &\cellcolor[HTML]{7A7AF2}14.09 \\\hline
\end{tabularx}
\egroup
\vspace{-2.3mm}
\caption{Evaluations on the Tasty dataset for our visual and text model along with comparison against video captioning ~\cite{venugopalan2015,zhou2018end}. Performance drops when the amount of pre-training decreases. Our method performs better than video captioning.} 
\label{fig:tasty_res}
\vspace{-0.4cm}
\end{table}

\vspace{-0.6cm}
\subsection{Supervised vs. Zero-Shot Learning}
\vspace{-0.15cm}
We compare the differences of supervised and zero-shot learning on YouCookII. We divide the dataset into four splits based on the 89 dishes and use three splits for training and half of the videos in the fourth split for testing. In the zero-shot setting, the videos from the other half of the fourth split are unused, while in the supervised setting, they are included as part of the training. We report results averaged over the four cross-folds in Table~\ref{fig:YoucookII_res_zero_shot}.

As expected, the predictions are better when the model is trained under a supervised setting in comparison to zero-shot. This is true for all inputs, with the same drop as observed previously when moving from text to video and when moving from ground truth video segments to fixed window segments. However, the difference between the supervised vs. zero-shot  (see Table~\ref{fig:YoucookII_res_zero_shot} ``Sup. Visual'' vs.\ ``Zero Visual'') is surprisingly much smaller than the difference between a supervised setting with and without pre-training on Recipe1M (``Sup. Visual'' vs.\ ``Sup. Visual no pre-train''). This suggests that having a large corpus for pre-training is more useful than repeated observations for a specific dish. 

While the test set of Tasty Videos is fully zero-shot, 183 videos are of recipes which occur with some variations in the training, while 72 are without any variations. As expected, when comparing the predictions on these subsets separately (see Table~\ref{fig:tasty_res_zero}), we observe higher performance on videos with variations, especially for the very difficult BLEU4 score. This suggest that our method generalizes better when it receives visually similar recipes. 
 
\begin{table}[ht]
\footnotesize
\bgroup
\def\arraystretch{1.25}
\setlength\tabcolsep{0.675em}
\setstretch{0.2}
\begin{tabularx}{0.99\columnwidth}{|c|r|r|r|r|r|}
\hline 
\scriptsize{Method} & \tiny{ING} & \tiny{VERBS} & \tiny{BLEU1} & \tiny{BLEU4} & \tiny{METEOR} \\ \hline 
Sup.~Visual (GT) &\cellcolor[HTML]{FF6CBC}20.93 &\cellcolor[HTML]{9DE79D}24.76 &\cellcolor[HTML]{FFC284}22.11 &\cellcolor[HTML]{FF8D8D}1.21 &\cellcolor[HTML]{6E6EF0}10.66 \\ 
Sup.~Visual &\cellcolor[HTML]{FF7CC3}18.90 &\cellcolor[HTML]{BBEEBB}23.15 &\cellcolor[HTML]{FFCC9A}21.09 &\cellcolor[HTML]{FF9E9E}1.03 &\cellcolor[HTML]{7979F2}10.22 \\ 
Sup.~Visual no pre-train &\cellcolor[HTML]{FFE9F5}2.69 &\cellcolor[HTML]{E9FAE9}19.43 &\cellcolor[HTML]{FFF5EA}15.05 &\cellcolor[HTML]{FFE5E5}0.15 &\cellcolor[HTML]{DCDCFB}5.89 \\ 
Sup.~Text &\cellcolor[HTML]{FF4FAE}24.56 &\cellcolor[HTML]{65DA65}27.24 &\cellcolor[HTML]{FF9F40}24.94 &\cellcolor[HTML]{FF4040}1.99 &\cellcolor[HTML]{4040EC}12.50 \\ \hline
Zero Visual (GT) &\cellcolor[HTML]{FF85C7}17.77 &\cellcolor[HTML]{BCEFBC}23.11 &\cellcolor[HTML]{FFD1A3}20.61 &\cellcolor[HTML]{FFAFAF}0.84 &\cellcolor[HTML]{8A8AF3}9.51 \\ 
Zero Visual &\cellcolor[HTML]{FFD7ED}6.04 &\cellcolor[HTML]{BAEEBA}23.19 &\cellcolor[HTML]{FFD4A9}20.30 &\cellcolor[HTML]{FFB6B6}0.76 &\cellcolor[HTML]{9090F4}9.27 \\ 
Zero Visual no pre-train &\cellcolor[HTML]{FFEDF7}1.58 &\cellcolor[HTML]{F0FBF0}17.83 &\cellcolor[HTML]{FFF5EC}14.54 &\cellcolor[HTML]{FFECEC}0.01 &\cellcolor[HTML]{ECECFD}5.03 \\ 
Zero Text &\cellcolor[HTML]{FF75BF}19.90 &\cellcolor[HTML]{9BE79B}24.86 &\cellcolor[HTML]{FFB76F}23.06 &\cellcolor[HTML]{FF7474}1.47 &\cellcolor[HTML]{6666F0}10.98 \\ \hline
\end{tabularx}
\egroup
\vspace{-2.0mm}
\caption{Comparison of zero-shot vs. supervised setting~(Sup.), on YouCookII~\cite{zhou2018towards} by cross validation. Supervised results are better overall. Without pre-training the performance drop is significant. }
\label{fig:YoucookII_res_zero_shot}
\vspace{-0.18cm}
\end{table}

\begin{table}[ht]
\footnotesize
\bgroup
\def\arraystretch{1.25}
\setlength\tabcolsep{1.005em}
\setstretch{0.5}
\begin{tabularx}{0.99\columnwidth}{|c|r|r|r|r|r|}
\hline 
\scriptsize{Method} & \tiny{ING} & \tiny{VERBS} & \tiny{BLEU1} & \tiny{BLEU4} & \tiny{METEOR} \\ \hline 
w/o variations &\cellcolor[HTML]{FFD8ED}14.20 &\cellcolor[HTML]{DEF7DE}17.08 &\cellcolor[HTML]{FFDCB8}16.67 &\cellcolor[HTML]{FFDBDB}0.76 &\cellcolor[HTML]{BDBDF8}10.00 \\ \hline
w/ variations &\cellcolor[HTML]{FF72BE}25.40 &\cellcolor[HTML]{84E184}20.41 &\cellcolor[HTML]{FFB366}20.54 &\cellcolor[HTML]{FF6666}2.16 &\cellcolor[HTML]{6666F0}13.00 \\ \hline
all videos &\cellcolor[HTML]{FF91CC}22.24 &\cellcolor[HTML]{A3E8A3}19.47 &\cellcolor[HTML]{FFBF80}19.45 &\cellcolor[HTML]{FF8A8A}1.77 &\cellcolor[HTML]{7F7FF2}12.15 \\ \hline
\end{tabularx}
\egroup
\vspace{-0.2cm}
\caption{Evaluations on the Tasty test set on videos with and without variations in the training set.We do better on variations.} 
\label{fig:tasty_res_zero}
\vspace{-0.4cm}
\end{table}

\vspace{-0.1cm}
\subsection{Knowledge Transfer}
\vspace{-0.1cm}
At the core of our method is the transfer of knowledge from text resources to solve a challenging visual problem. We evaluate the effectiveness of the knowledge transfer by varying the amount of training data from Recipe1M to be used for pre-training. Looking at the averaged scores over all the predicted steps on Tasty Videos, we observe a decrease in all evaluation measures as we limit the amount of data from Recipe1M (see Table~\ref{fig:tasty_res}, ``Ours Text'' 100\%, 50\%, etc.), with the most significant decrease occuring for the BLEU4 score. If there is no pre-training, \ie when the model learned only on text from Tasty Videos (``Ours Text (0\%)''), the decrease in scores is noticeable for all evaluation criteria. These results again verify that pre-training has a significant effect on our method's performance.

\vspace{-0.05cm}
\subsection{Comparisons to Video Captioning} 

We compare our method against different video captioning methods in Tables~\ref{fig:tasty_res} and ~\ref{fig:YoucookII_res_supervised} for Tasty Videos and YouCookII respectively. Unlike predicting future steps, captioning methods generate sentences after observing their visual data which makes it a much easier task than future prediction. We train and test S2VT~\cite{venugopalan2015}, an RNN based encoder-decoder approach, on the ground truth segments of the Tasty dataset. Our visual model outperforms this baseline, especially for ingredient recall, by 13\%, and with an improvement of 0.3 in BLEU4 score in Table~\ref{fig:tasty_res}. To highlight the difficulty of predicting future steps compared to captioning, we train S2VT~\cite{venugopalan2015} for predicting the next step from the observation of the current step (see Table~\ref{fig:tasty_res} ``S2VT~\cite{venugopalan2015} next (GT)''). Our visual model outperforms this variation with a big margin for all scores. We also tested the End-to-end Masked Transformer~\cite{zhou2018end} on our dataset and get a BLEU4/METEOR of 0.54 / 5.48 (vs. our 1.23 / 11.00). The poor performance is likely due to the increased dish diversity and difficulty of our dataset vs YouCook2. 

We compare our model on the validation set of the YouCookII dataset against two state-of-the-art video captioning methods~\cite{yao2015describing,zhou2018end} in Table~\ref{fig:YoucookII_res_supervised}. End-to-end Masked Transformer~\cite{zhou2018end} performs dense video captioning by both localizing steps and generating descriptions for these steps, while TempoAttn~\cite{yao2015describing} is an RNN-based encoder-decoder approach. Again, even though our task is more difficult than captioning, our method outperforms both of the captioning methods in BLEU4 and METEOR scores. Compared to the state-of-the-art~\cite{zhou2018end}, our visual model achieves a METEOR score that is twice as high and a BLEU4 score four times higher. We attribute the better performance of our method compared to the captioning methods to the pre-training on the Recipe1M dataset which allows our model to generalize. Note that for YouCookII, as we use all the videos in the training set, our training is no longer a zero-shot but a supervised scenario.

 \begin{table}[ht]
\footnotesize
\bgroup
\def\arraystretch{1.25} 
\setlength\tabcolsep{0.535em}
\setstretch{0.3}
\begin{tabularx}{0.99\columnwidth}{|c|r|r|r|r|r|}
\hline 
\scriptsize{Method} & \scriptsize{ING} & \scriptsize{VERBS} & \scriptsize{BLEU1} & \scriptsize{BLEU4} & \scriptsize{METEOR} \\ \hline 
TempoAttn(GT)~\cite{yao2015describing} & - & - & - &\cellcolor[HTML]{FFCBCB}0.87 &\cellcolor[HTML]{ACACF7}8.15 \\ 
End-to-end(GT)~\cite{zhou2018end} & - & - & - &\cellcolor[HTML]{FFA7A7}1.42 &\cellcolor[HTML]{6D6DF0}11.20 \\ 
Ours Visual (GT) &\cellcolor[HTML]{FF6EBC}21.36 &\cellcolor[HTML]{78DE78}27.55 &\cellcolor[HTML]{FFB368}23.71 &\cellcolor[HTML]{FF9595}1.66 &\cellcolor[HTML]{6666F0}11.54 \\ \hline
TempoAttn~\cite{yao2015describing} & - & - & - &\cellcolor[HTML]{FFECEC}0.08 &\cellcolor[HTML]{ECECFD}4.62 \\ 
End-to-end~\cite{zhou2018end} & - & - & - &\cellcolor[HTML]{FFE6E6}0.30 &\cellcolor[HTML]{CACAFA}6.58 \\ 
Ours Visual &\cellcolor[HTML]{FF90CC}17.64 &\cellcolor[HTML]{90E490}25.11 &\cellcolor[HTML]{FFBB77}22.55 &\cellcolor[HTML]{FFAAAA}1.38 &\cellcolor[HTML]{7777F1}10.71 \\ \hline
Ours Text &\cellcolor[HTML]{FF4FAE}24.60 &\cellcolor[HTML]{65DA65}29.39 &\cellcolor[HTML]{FF9F40}26.49 &\cellcolor[HTML]{FF4040}2.66 &\cellcolor[HTML]{4040EC}13.31 \\ \hline
\end{tabularx}
\egroup
\vspace{-2.5mm}
\caption{Comparison against captioning methods on the YouCookII~\cite{zhou2018towards} validation set. We perform better than the state-of-the-art captioning methods.} 
\label{fig:YoucookII_res_supervised}
\vspace{-0.3cm}
\end{table}
 
\vspace{-0.07cm}
\subsection{Human Ratings}\label{sec:human_study} 
\begin{figure}[ht]
\centering 
\includegraphics[width=0.99\columnwidth]{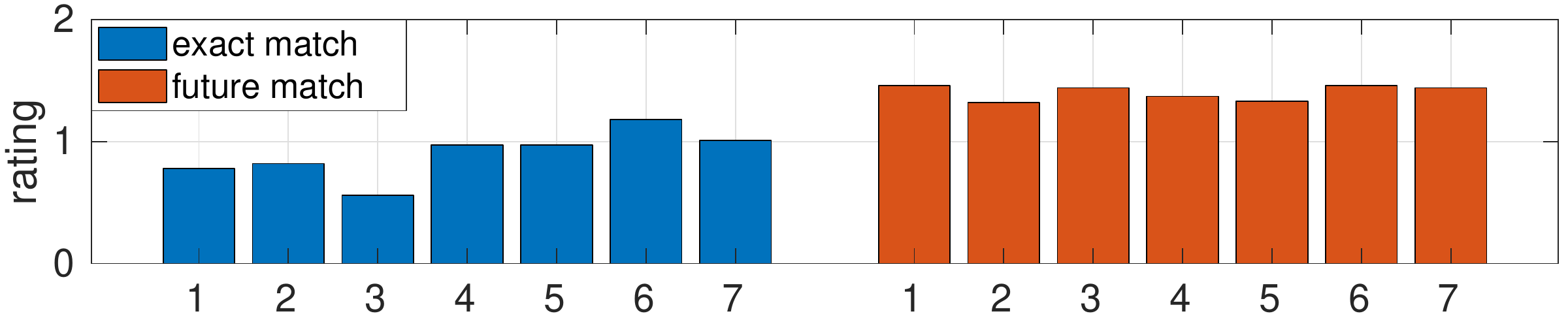} 
\vspace{0.5em} 
\includegraphics[width=0.99\columnwidth]{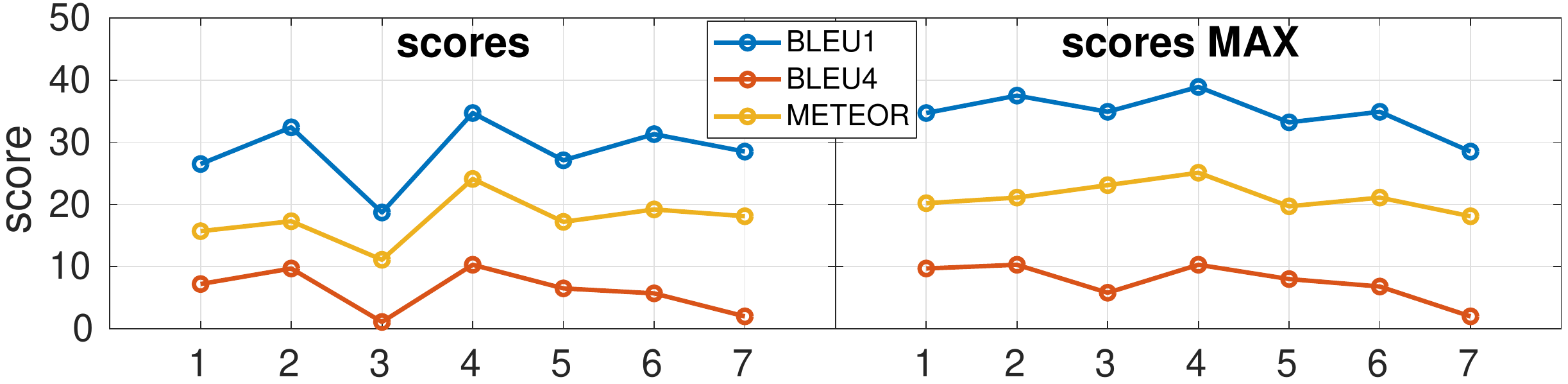} 
 \vspace{-5.0mm}
\caption{Comparison of human ratings (upper graph) versus automated sentence scores (lower graph) over a subset of 30 recipes.}
\vspace{-2.5mm}
\label{fig:human_selected} 
\vspace{-0.3cm}
\end{figure}

\vspace{-0.07cm}
We ask human raters to directly assess how well the predicted steps match the ground truth with scores 0 (`not at all'), 1 (`somewhat') or 2 (`very well'). If the prediction receives a score of 0, 
we additionally ask the human to judge if the predicted step is still a plausible future prediction, again with the same scores of 0 (`not at all'), 1 (`somewhat'), or 2 (`very likely').
We conduct this study with 3 people on a subset of 30 recipes from the test set, each with 7 steps, and present their ratings in 
Fig.~\ref{fig:human_selected} while comparing them to automated sentence scores.

In Fig.~\ref{fig:human_selected}, the upper graph shows the results of the human raters and the lower graph shows the automated sentence scores. Raters report a score close to 1 for the initial step predictions indicating that our method, even by only seeing the ingredients, can start predicting plausible steps. Scores increase towards the end of the recipe and are lowest at step 3. The average score of the predicted steps being a possible future prediction are consistently high across all steps. Even if the predicted step does not exactly match the ground truth, human raters still consider it possible for the future, including the previously low rating for step 3. Overall, the ratings indicate that the predicted steps are plausible.

The lower graph in Fig.~\ref{fig:human_selected} shows automated scores for the same user study recipes. 
The left plot shows the standard scores for the predicted sentences matching the ground truth; overall, trends are very similar to the user study, including the low-scoring step 3. To match the second setting of the user study, we compute the sentence scores between the predicted sentence $\hat{s}_j$ and all future ground truth steps $\{s_j, s_{j+1}, s_{j+2},s_{j+3}\}$ and select the step with the maximum score as our future match. These scores are plotted in the lower right of Fig.~\ref{fig:human_selected}; similar to the human study, sentence scores increase overall. 

\vspace{-0.1cm}
\subsection{Ablation Study} 
\vspace{-0.1cm}
Since our method is modular, we conduct an ablation study to check the interchangeability of the sentence encoder on the Recipe1M dataset~\cite{salvador2017learning}. 
Instead of using our own sentence encoder, we represent the sentences using ST vectors trained on the Recipe1M dataset, as provided by the authors~\cite{salvador2017learning}. These vectors have been shown to perform well for their recipe retrieval. Our results, presented in the supplementary text show that our sentence encoder performs on par with ST encodings. However, our encoder, model and decoder can all be trained jointly and do not require a separate pre-training of a sentence autoencoder.

\vspace{-0.2cm}
\section{Conclusion}
\vspace{-0.2cm}
In this paper we present a method for zero-shot action anticipation in videos. Our model learns to generalize instructional knowledge from the text domain. Applying this knowledge to videos allows us to tackle the challenging task of predicting steps of complex tasks from visual data, which is otherwise ruled out because of scarcity of or difficulty to annotate training data. We present a new, diverse dataset of cooking videos, which is of high interest for the community. We successfully validate our method's performance on both text and video data. We show that our model is able to produce coherent and plausible future steps. We conclude that our knowledge transfer strategy works much better than captioning methods and generalizes well on different datasets. In the future we hope to include more information into our model, such as the title of the recipe.  
 
\vspace{-0.5cm} 
\paragraph{Acknowledgments} This work has been partly funded by the Deutsche Forschungsgemeinschaft (DFG, German Research Foundation) YA 447/2-1 and GA 1927/4-1 (FOR 2535 Anticipating Human Behavior) and partly by the Singapore Ministry of Education Academic Research Fund Tier 1. We thank Sven Behnke and Juergen Gall for useful discussions.

{\small
\bibliographystyle{ieee}
\bibliography{ms}
}

\end{document}

%%%%%%%%% TITLE
\title{Supplementary: Zero-Shot Anticipation for Instructional Activities}

\author{Fadime Sener\\
University of Bonn, Germany\\ 
{\tt\small sener@cs.uni-bonn.de}
\and
Angela Yao\\
National University of Singapore\\ 
{\tt\small ayao@comp.nus.edu.sg}
}

\maketitle

\section{Tasty Videos Dataset}

\begin{figure}[ht]
\centering 
\includegraphics[width=\columnwidth]{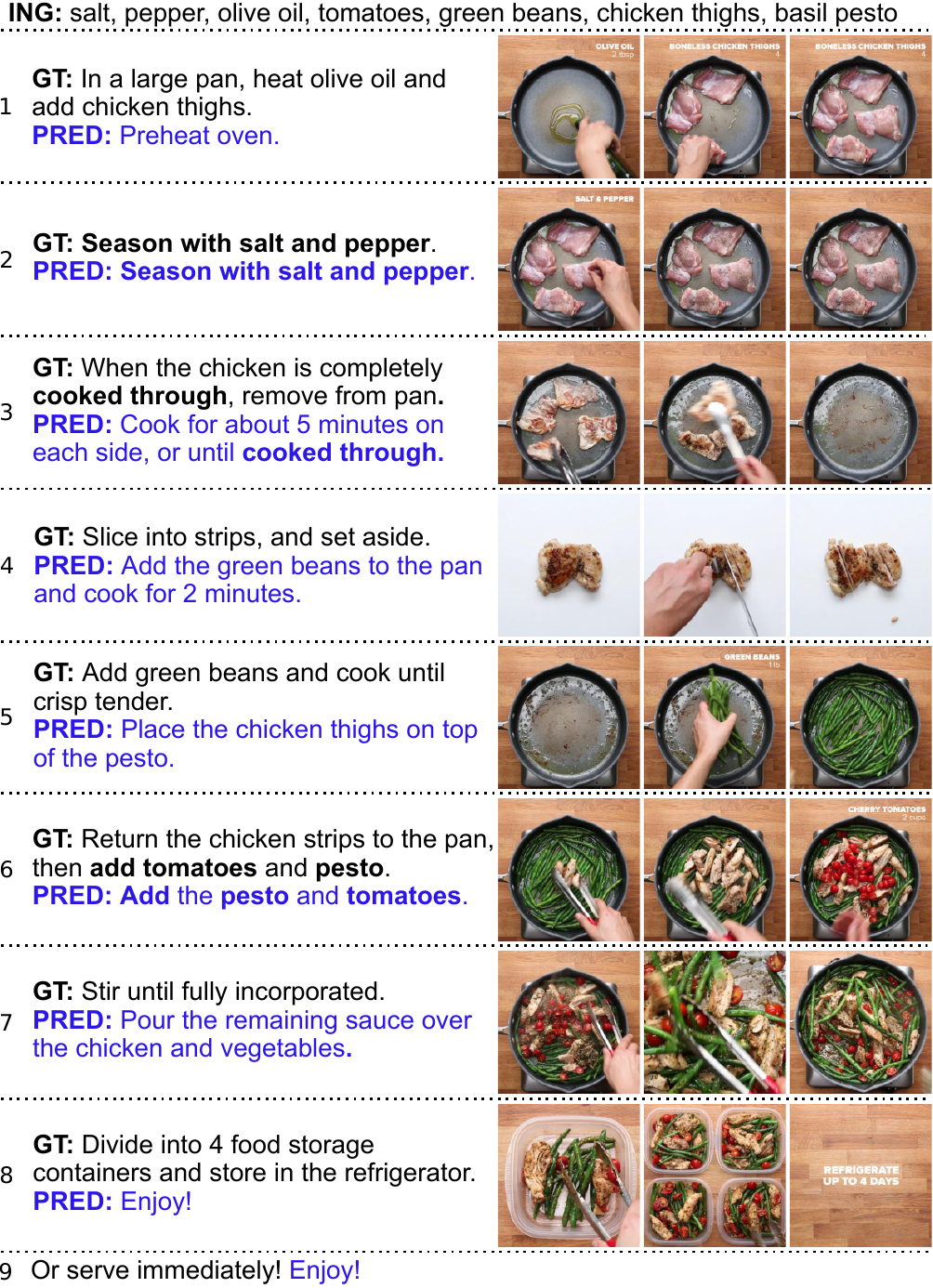}
\vspace{-1.8mm}
\caption{Next step prediction of our visual model for ``Weekday Meal-prep Pesto Chicken and Veggies''. The blue sentences are our model's predictions. Our model's prediction for step4 is a plausible future step prediction as it happens in step5. Our model's predictions for step8 and step9 are plausible recommendations. Note that our model predicts the next steps before seeing these segments!}
\vspace{-0.3cm}
\label{fig:videoexp_Weekday} 
\end{figure}

The videos in the Tasty Videos Dataset are captured with a fixed overhead camera and are focused entirely on the preparation of the dish. Videos are designed to be primarily visually informative without any narrations, except for the textual recipe steps. An example video for ``Weekday Meal-prep Pesto Chicken and Veggies'' can be found online here: \texttt{\small\url{https://tasty.co/recipe/weekday-meal-prep-pesto-chicken-veggies}}, which is shown in Fig.~\ref{fig:videoexp_Weekday}. More videos can be found on \texttt{\small\url{https://tasty.co/}}.

\begin{figure}[ht]
\centering 
\includegraphics[width=0.85\columnwidth]{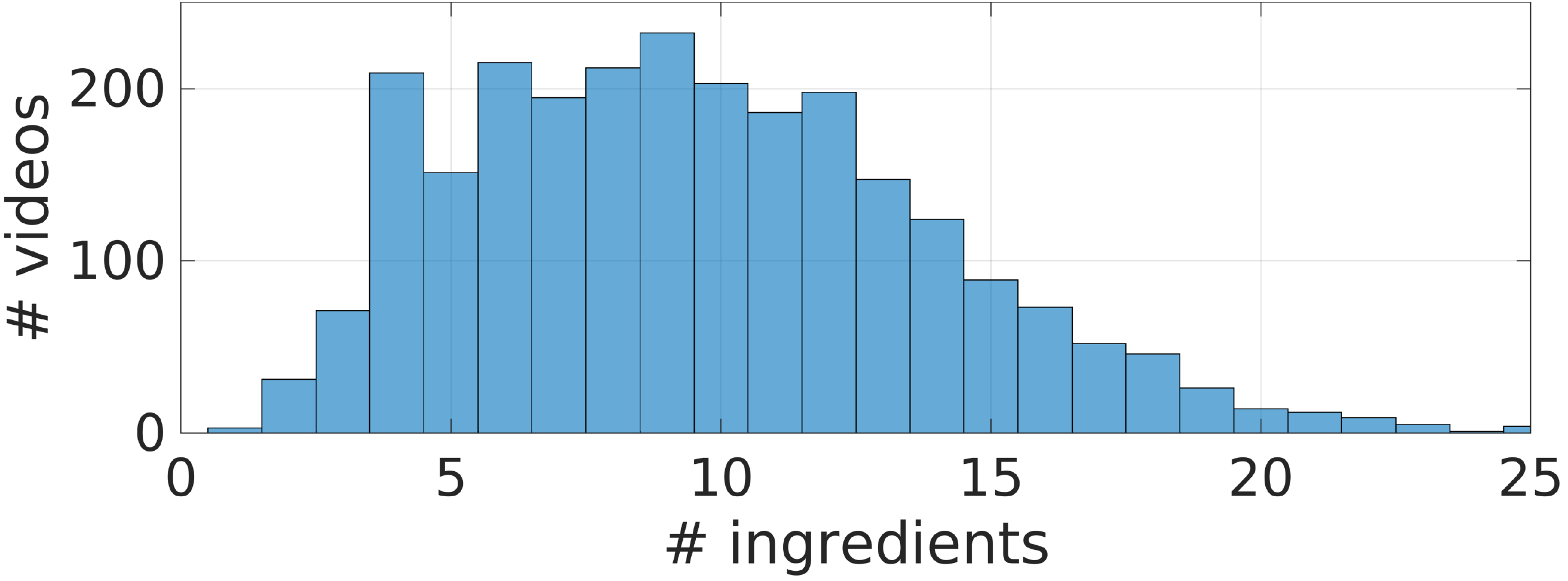} 
\vspace{-0.3cm}
\caption{Distribution of the number of ingredients (out of $1199$ unique ingredients).}
\vspace{-0.3cm}
\label{fig:tasty_ingredients} 
\end{figure} 

\begin{figure}[htb]
\centering 
\includegraphics[width=0.65\columnwidth,height=0.45\columnwidth]{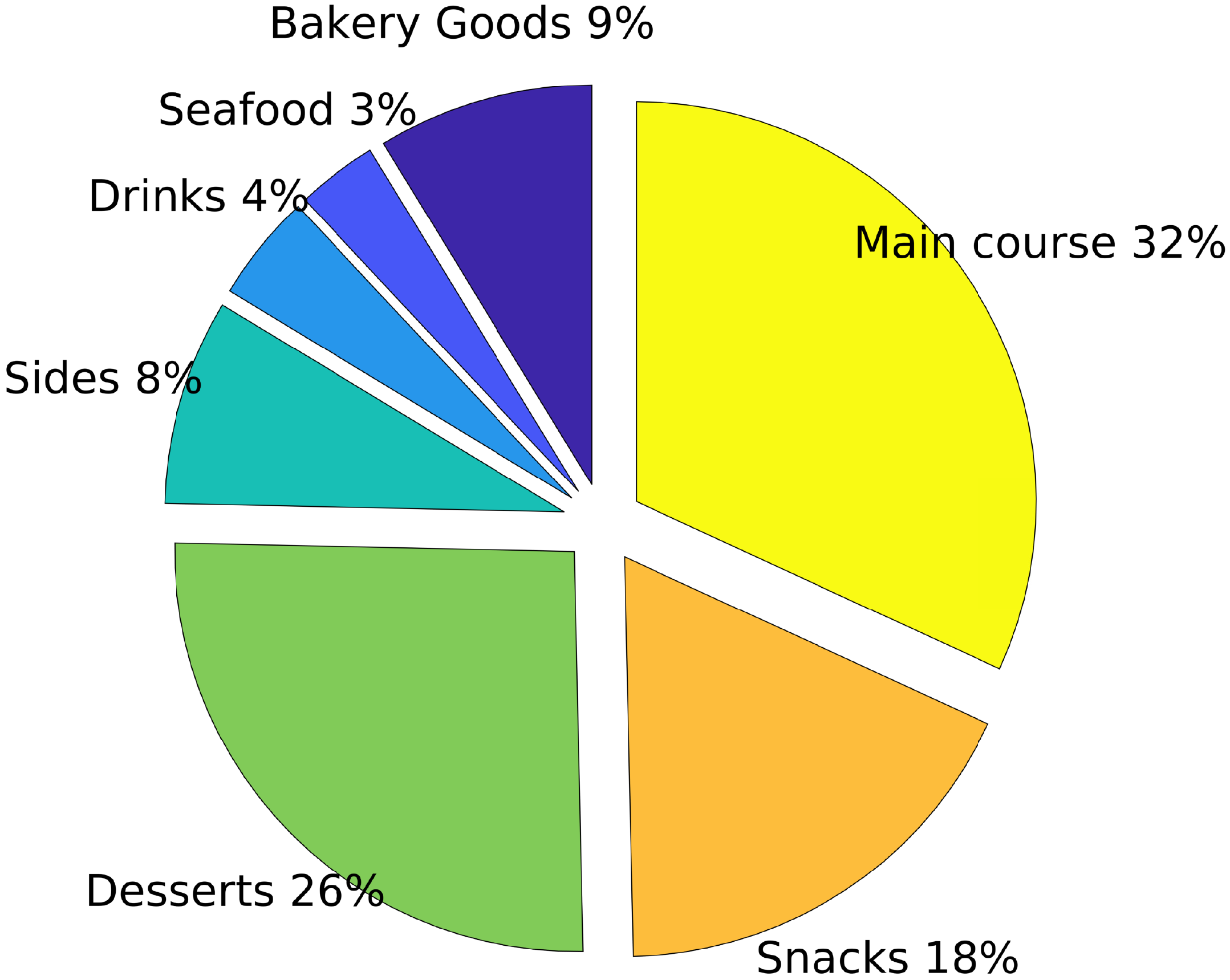} 
\vspace{-0.2cm}
\caption{Rough categorization of the recipes in our dataset.} 
\vspace{-0.3cm}
\label{fig:tasty_pie} 
\end{figure} 

In our dataset, there are $1199$ unique ingredients and the average number of ingredients is 9, see Figure~\ref{fig:tasty_ingredients}. In comparison, the number of unique ingredients in the Recipe1M dataset~\cite{salvador2017learning} is around 4K. 
Our dataset has a large variety of meals, including main courses, snacks, sides etc., see Fig.~\ref{fig:tasty_pie}. 
 
\begin{figure}[htb]
\centering 
\includegraphics[width=0.85\columnwidth]{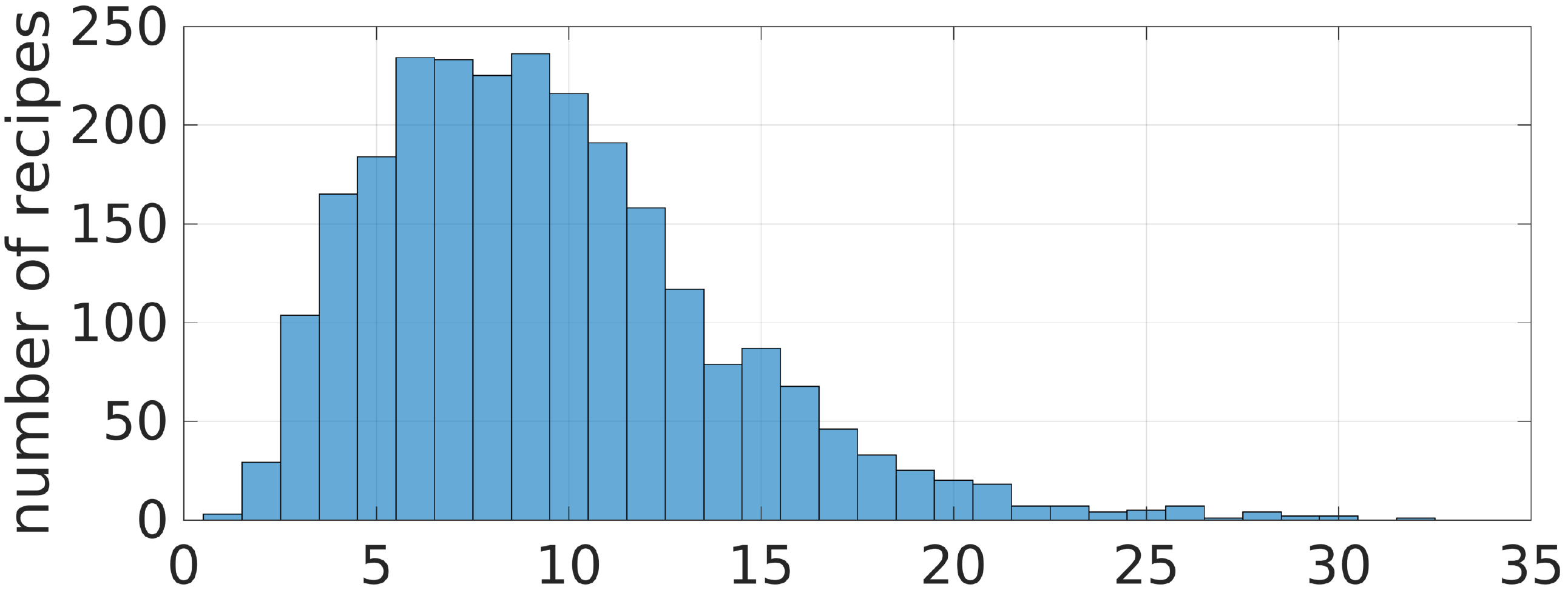} 
\vspace{-0.2cm}
\caption{Distribution of the number of visual steps. The average number is 9.}
\vspace{-0.3cm}
\label{fig:tasty_steps} 
\end{figure} 

\begin{figure}[htb]
\centering 
\includegraphics[width=0.85\columnwidth]{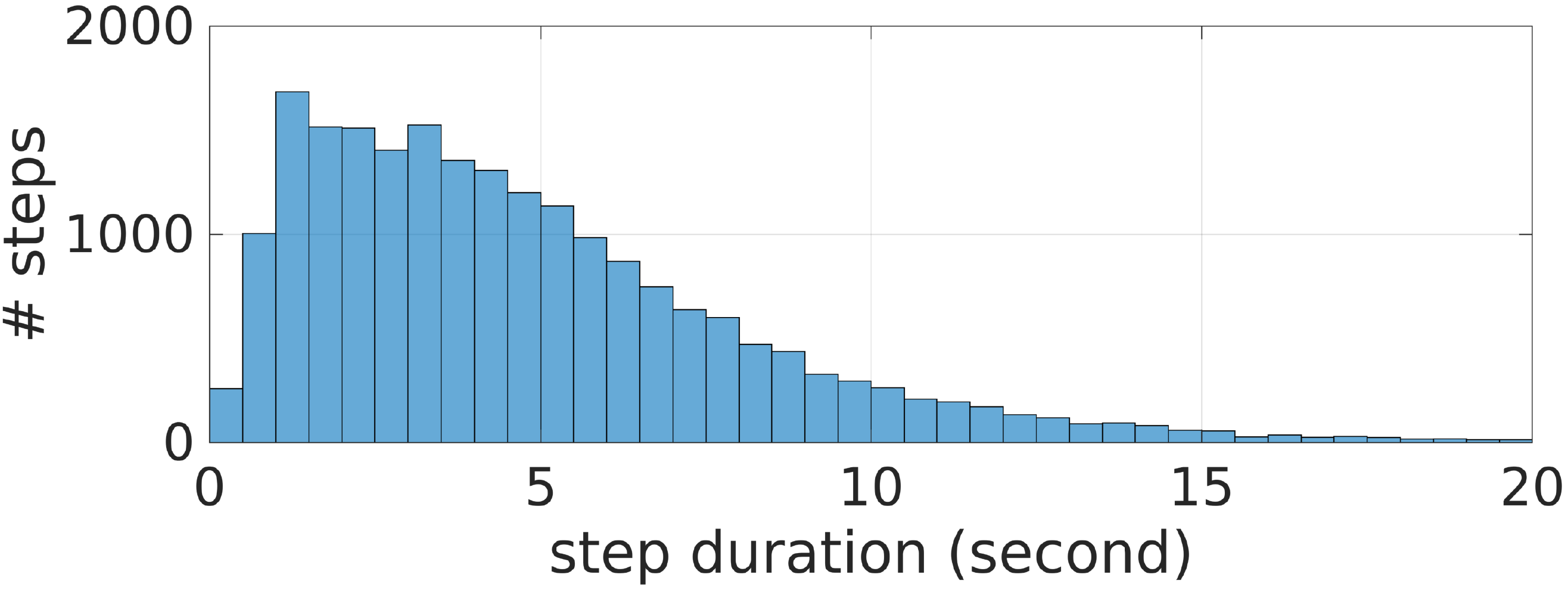} 
\vspace{-0.2cm}
\caption{Distribution of annotated visual step durations.}
\vspace{-0.3cm}
\label{fig:tasty_durations_step} 
\end{figure}

\begin{figure}[htb]
\centering 
\includegraphics[width=0.85\columnwidth]{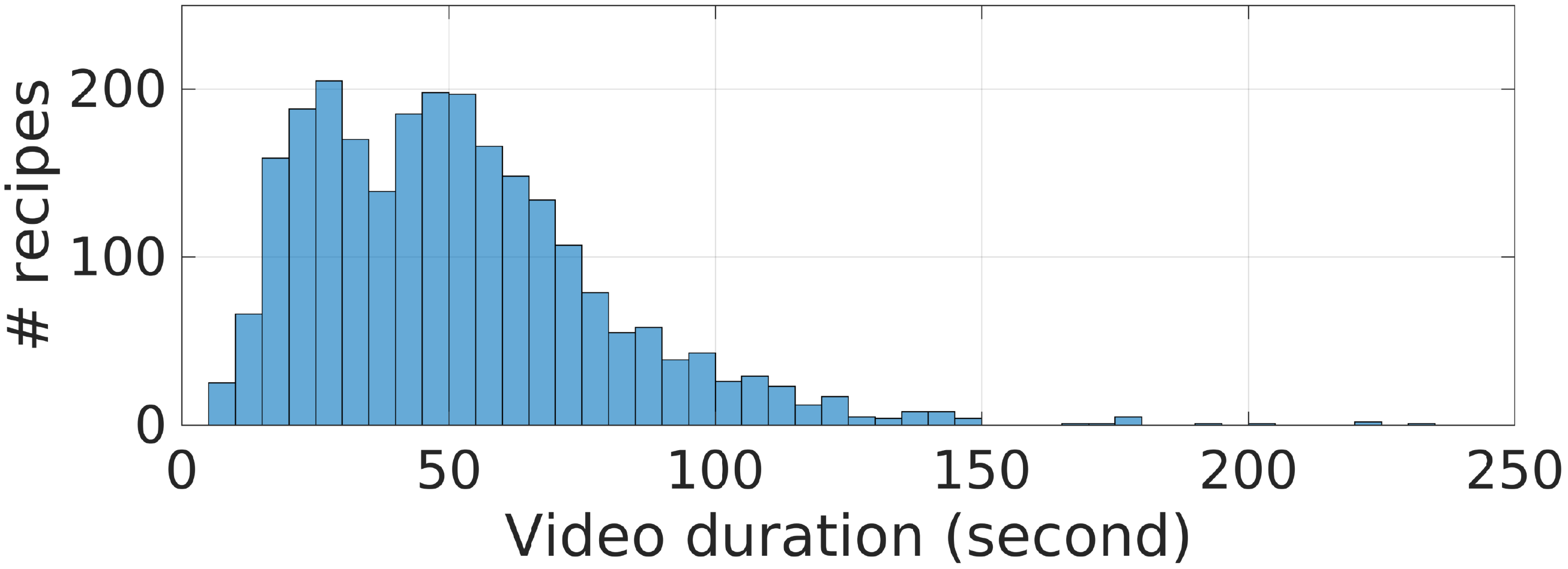} 
\vspace{-0.2cm}
\caption{Distribution of video durations.}
\vspace{-0.3cm}
\label{fig:tasty_durations} 
\end{figure}

Each recipe has a list of instructions. For each recipe step, we annotate the temporal boundaries in which the step occurs within the video, omitting those without visual correspondences, such as alternative recommendations. As such, there are less visual instructions than text-based ones.

The average number of visual recipe steps is 9, and there are $21236$ visual recipe steps in total. In Figure~\ref{fig:tasty_steps}, we show the distribution of the number of visual recipe steps.  In Figure~\ref{fig:tasty_durations_step}, we report the distribution of the duration of the annotated visual steps. The average visual step duration is 144 frames or 5 seconds. In Figure~\ref{fig:tasty_durations}, we report the distribution of the duration of our videos. The shortest video lasts 6 seconds while the longest lasts 233 seconds. The average video duration is 1551 frames or 54 seconds.

\begin{figure*}[htb]
\centering 
\includegraphics[width=\textwidth]{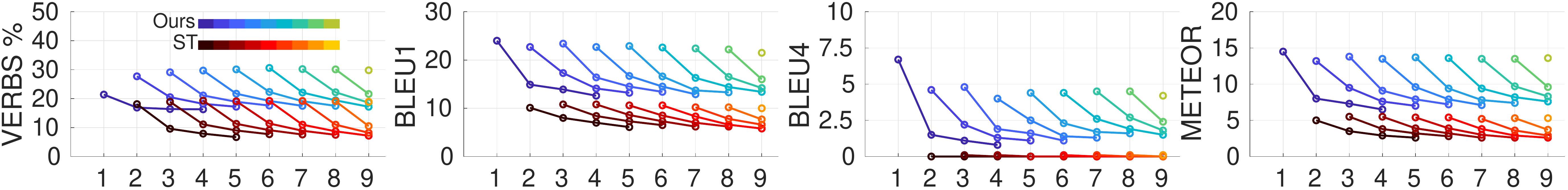}
\vspace{-5mm}
\caption{The recall of verbs and sentence scores computed between the predicted and ground truth sentences for our model (Ours) and the skip-thoughts (ST) model over the entire test set of the Recipe1M dataset~\cite{salvador2017learning}. The x-axes indicate the step number being predicted in the recipe; each curve begins on the first (relative) prediction, \ie the $(j+1){\text{th}}$ step after having received steps $1$ to $j$ as input.} 
\label{fig:sents_ours_all} 
\vspace{-0.1cm}
\end{figure*}

\begin{figure*}[htb]
 \centering
 \begin{subfigure}[b]{0.65\textwidth}
 \centering
 \includegraphics[width=\textwidth]{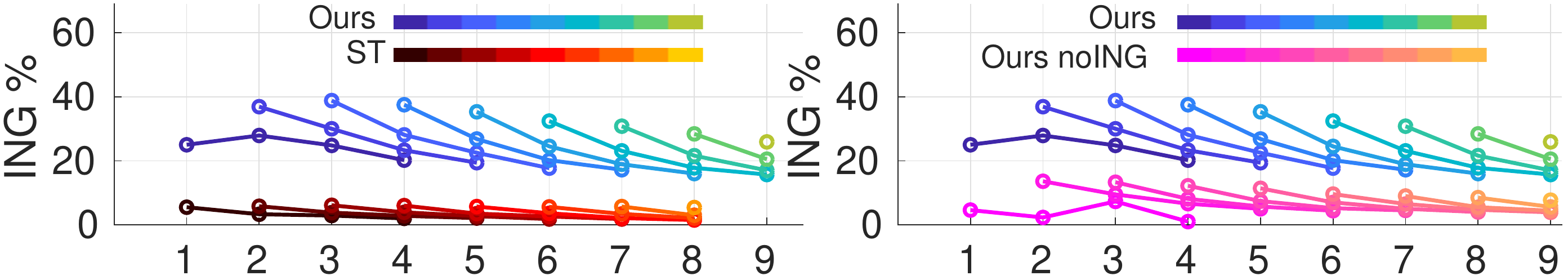} 
 \label{fig:inredients_ours_all}
 \end{subfigure}
 \hfill
 \centering
 \begin{subfigure}[b]{0.33\textwidth}
 \centering
 \includegraphics[width=\textwidth]{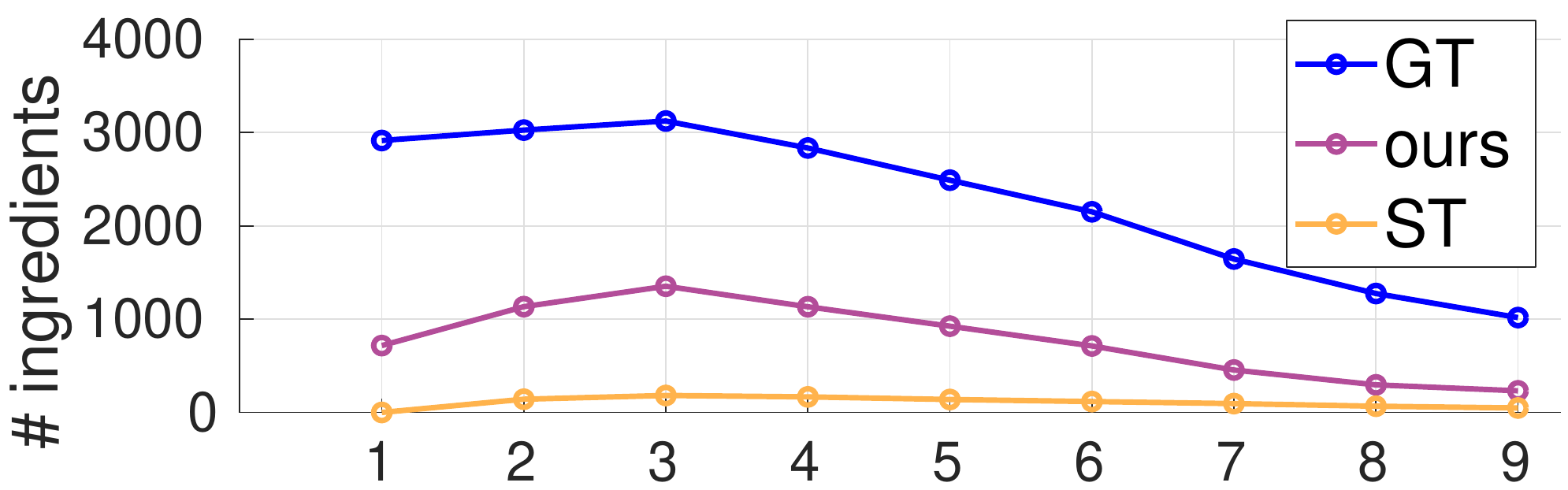} 
 \label{fig:ingredient_trend}
 \end{subfigure} 
\vspace{-5mm}
\caption{a) The recall of ingredients predicted by our model (Ours), by our model trained without the ingredients (Ours noING) and by skip-thoughts model (ST) over the entire test set of the Recipe1M dataset~\cite{salvador2017learning}. The x-axes in the plots indicate the step number being predicted in the recipe; each curve begins on the first (relative) prediction, \ie the $(j+1){\text{th}}$ step after having received steps $1$ to $j$ as input. b) Absolute number of ingredients detected in the ground truth steps (GT), steps predicted by our model (ours) and the skip-thoughts model (ST) computed over recipes with exactly 9 steps. The number of ingredients detected in a recipe decreases towards the end of the recipe.} 
\vspace{-0.1cm}
\label{fig:inredients_ours_all_ingredient_trend} 
\end{figure*}

\section{Experiments}

\subsection{Learning of Procedural Knowledge} 
In our experiments, we first target important keywords, specifically ingredients and verbs, since they indicate the next active object and next actions. 
Key ingredients and verbs alone do not capture the rich instructional nature of recipe steps, compare \eg \emph{`whisk'} and \emph{`egg'} to 
\emph{`Whisk the eggs till light and fluffy'}. As such, we also evaluate the predicted sentences as a whole and compare to ST predictions based on standard sentence evaluation metrics, such as BLEU (BiLingual Evaluation Understudy)~\cite{papineni2002bleu} and the METEOR score (Metric for Evaluation of Translation with Explicit ORdering)~\cite{banerjee2005meteor}. BLEU computes an n-gram based precision for predicted sentences with respect to ground truth sentences. METEOR creates an alignment between the ground truth and predicted sentence using the exact word matches, stems, synonyms, and paraphrases; it then computes a weighted F-score with an alignment fragmentation penalty. 
For the uninformed reader, we note that these scores are best at indicating precise word matches to ground truth. Yet in natural spoken language, much variation may exist between sentences conveying the same ideas. This is the case even in text with very specific language such as cooking recipes. For example, for the ground truth \emph{`Garnish with the remaining Wasabi and sliced green onions.'}, our method may predict \emph{`Transfer to a serving bowl and garnish with reserved scallions.'}. For a human reader, this is half correct, especially since \emph{`scallions'} and \emph{`green onions'} are synonyms, yet this example would have only a BLEU1 score of $30.0$, BLEU4 of $0.0$ and METEOR of $11.00$. 

We report our results over the entire test set of the Recipe1M dataset~\cite{salvador2017learning} in Figures~\ref{fig:sents_ours_all} (verbs and sentences),~\ref{fig:inredients_ours_all_ingredient_trend} (ingredients). We report scores of the predicted steps averaged over multiple recipes. Only those recipes which have at least $j$ steps contribute to the average for step $j$. Compared to the recipes with exactly 9 steps, results over the entire test set are not significantly different in trends. Based on the ground truth, we observe that the majority of the ingredients occurs in the early and middle steps and decreases in the last steps, see Figure~\ref{fig:inredients_ours_all_ingredient_trend}.

\subsection{Human study}
To assess the reliability of agreement between our human raters, we use Fleiss's kappa~\cite{fleiss1971mns} measure. It is used to analyze how much the annotators agree in their decisions. High level of agreement (at most 1) indicates that the human rating study was reliable.
Inter-rater agreement, measured via Fleiss's kappa~\cite{fleiss1971mns} by aggregating across all rating tasks, is 0.43, which is statistically significant at $p<0.05$. 
 
\begin{figure}[ht]
\centering 
\includegraphics[width=\columnwidth]{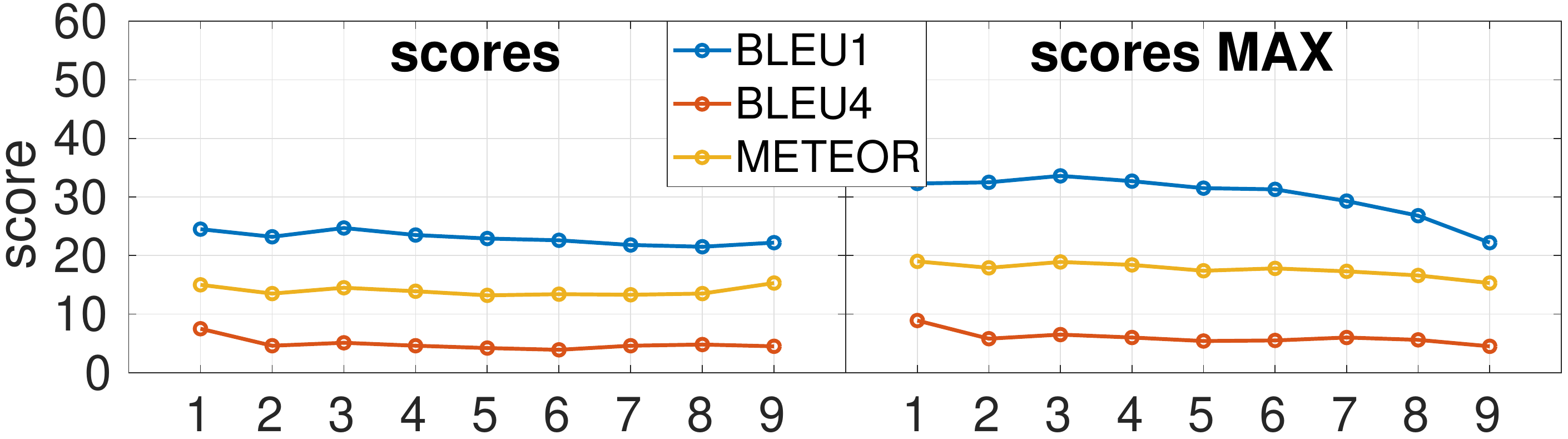}
\vspace{-4.5mm}
\caption{Comparison of the sentence scores versus the sentence scores computed based on the maximum match. Results reported over recipes with exactly 9 steps from Recipe1M dataset~\cite{salvador2017learning}. The x-axes indicate the step number being predicted in the recipe.}
\label{fig:sentence_max} 
\vspace{-0.3cm}
\end{figure} 

\begin{figure*}[ht]
\centering 
\includegraphics[width=\textwidth]{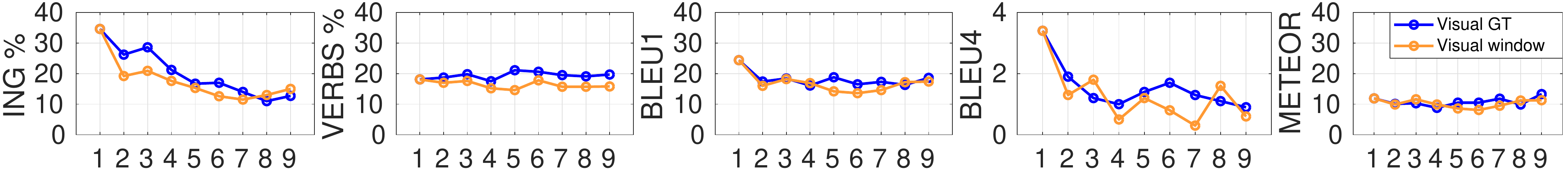}
\vspace{-5.5mm}
\caption{We compare the performance of our visual models for the recall of predicted ingredients and verbs, and sentence scores. Compared to using GT segments, the fixed windows show lower results but the results follow similar trends. } 
\label{fig:tasty_all_res_video} 
\vspace{-4mm}
\end{figure*}

In our human study, we observe that even if the predicted step does not exactly match the ground truth, human raters still consider it possible for the future. Following this setting of the user study, we compute the sentence scores between the predicted sentence $\hat{s}_j$ and all future ground truth steps $\{s_j, s_{j+1}, s_{j+2},s_{j+3}\}$ and select the step with the maximum score as our future match. We show the results for recipes with exactly 9 steps in Figure~\ref{fig:sentence_max}. The left plot shows the standard scores between the predicted sentences and the ground truth. The right plot shows the scores computed based on the maximum future match. 

We show some examples of predictions of our text-based method in Figures~\ref{fig:example_preds_Cheddar},\ref{fig:example_preds_Ambrosia},\ref{fig:example_preds_Fusilli},\ref{fig:example_preds_Baked},\ref{fig:example_preds_Green},\ref{fig:example_preds_Cheese},\ref{fig:example_preds_Slow},\ref{fig:example_preds_Summer}, along with the automated scores and human ratings.

\begin{figure*}[ht]
\centering 
\includegraphics[width=0.88\textwidth]{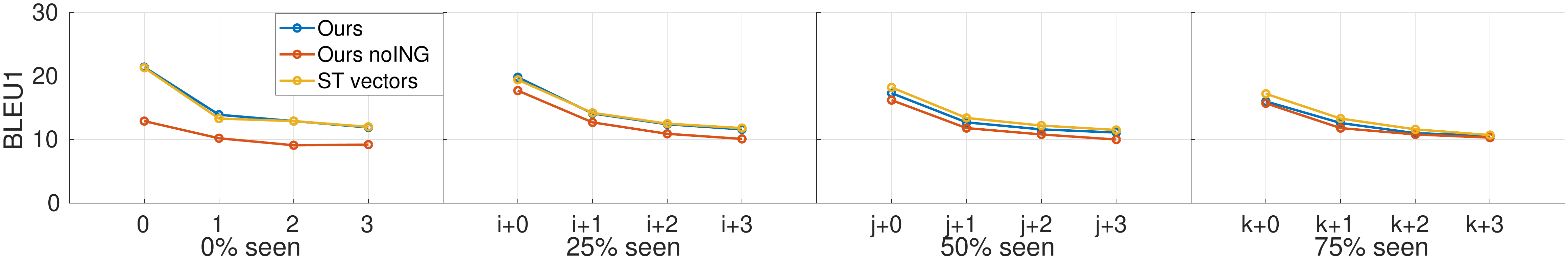} 
\includegraphics[width=0.88\textwidth]{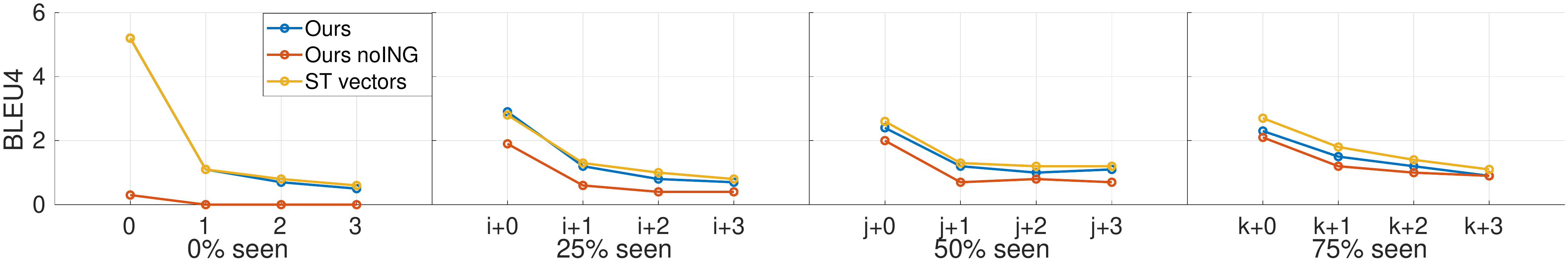} 
\includegraphics[width=0.88\textwidth]{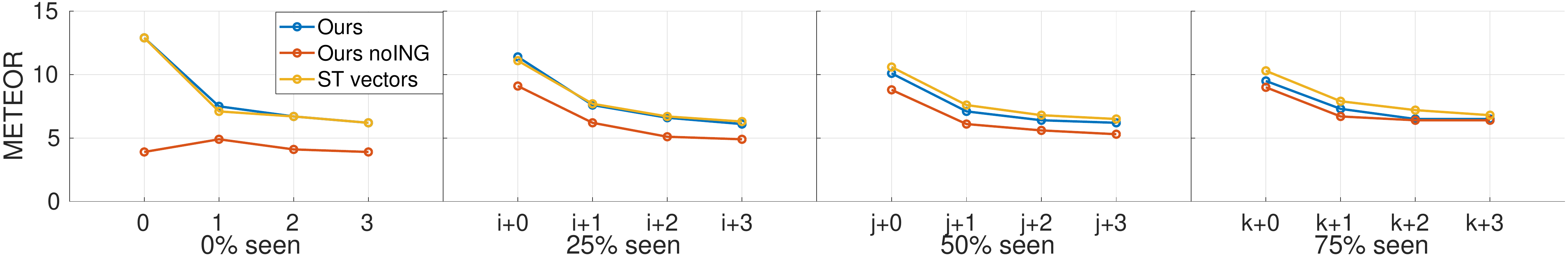} 
\vspace{-2.5mm}
\caption{Ablation study to check the interchangeability of the sentence encoder and how important the ingredients are as input for our method, computed over the entire test set of the Recipe1M dataset. ``$X\%$ seen'' refers to the number of steps the model receives as input, while predicting the remaining $(100 - X)\%$.}
\label{fig:im2recipe_percentage} 
\vspace{-3mm}
\end{figure*} 

\begin{figure*}[ht]
\centering 
\includegraphics[width=0.9\columnwidth]{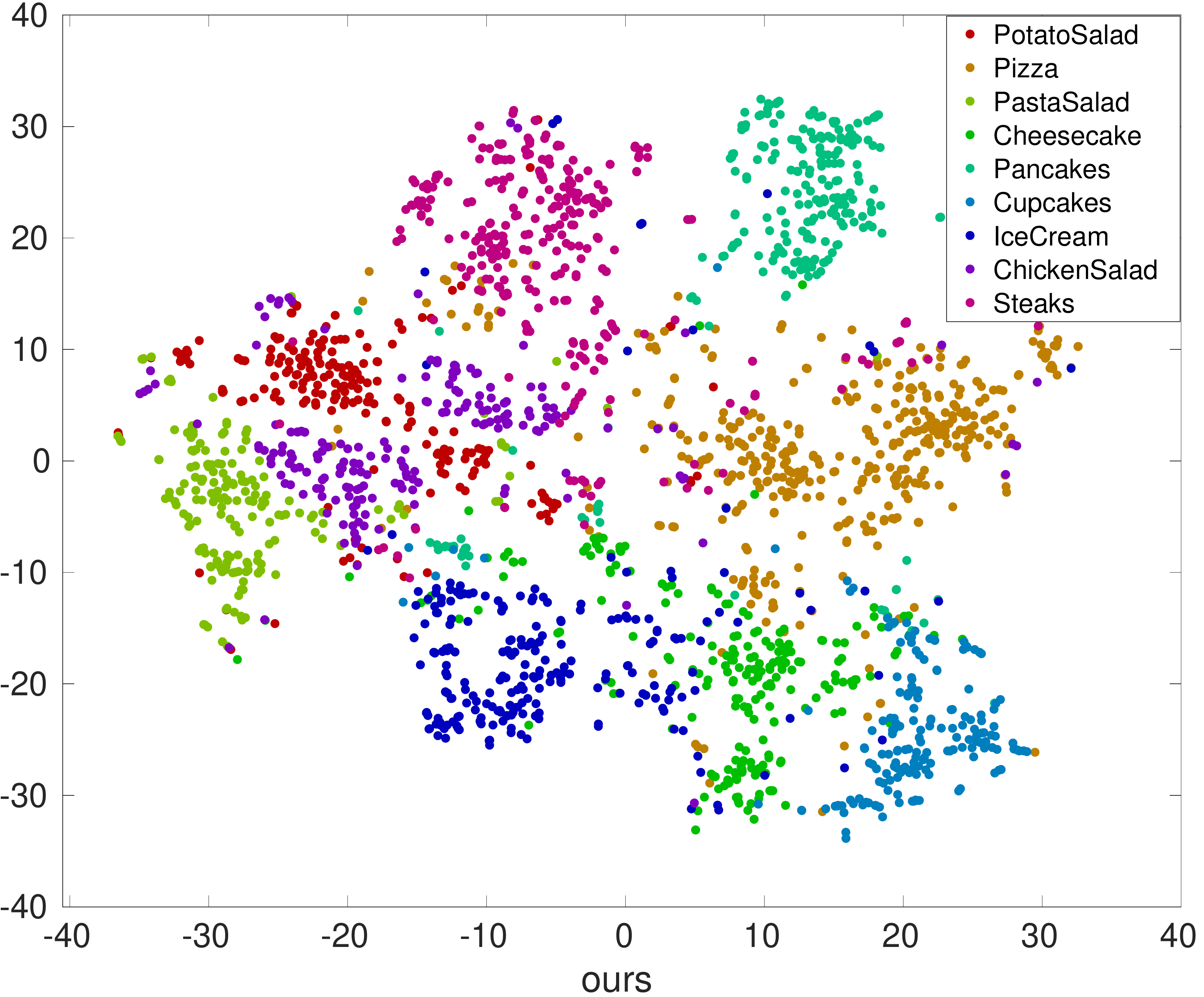}
\includegraphics[width=0.9\columnwidth]{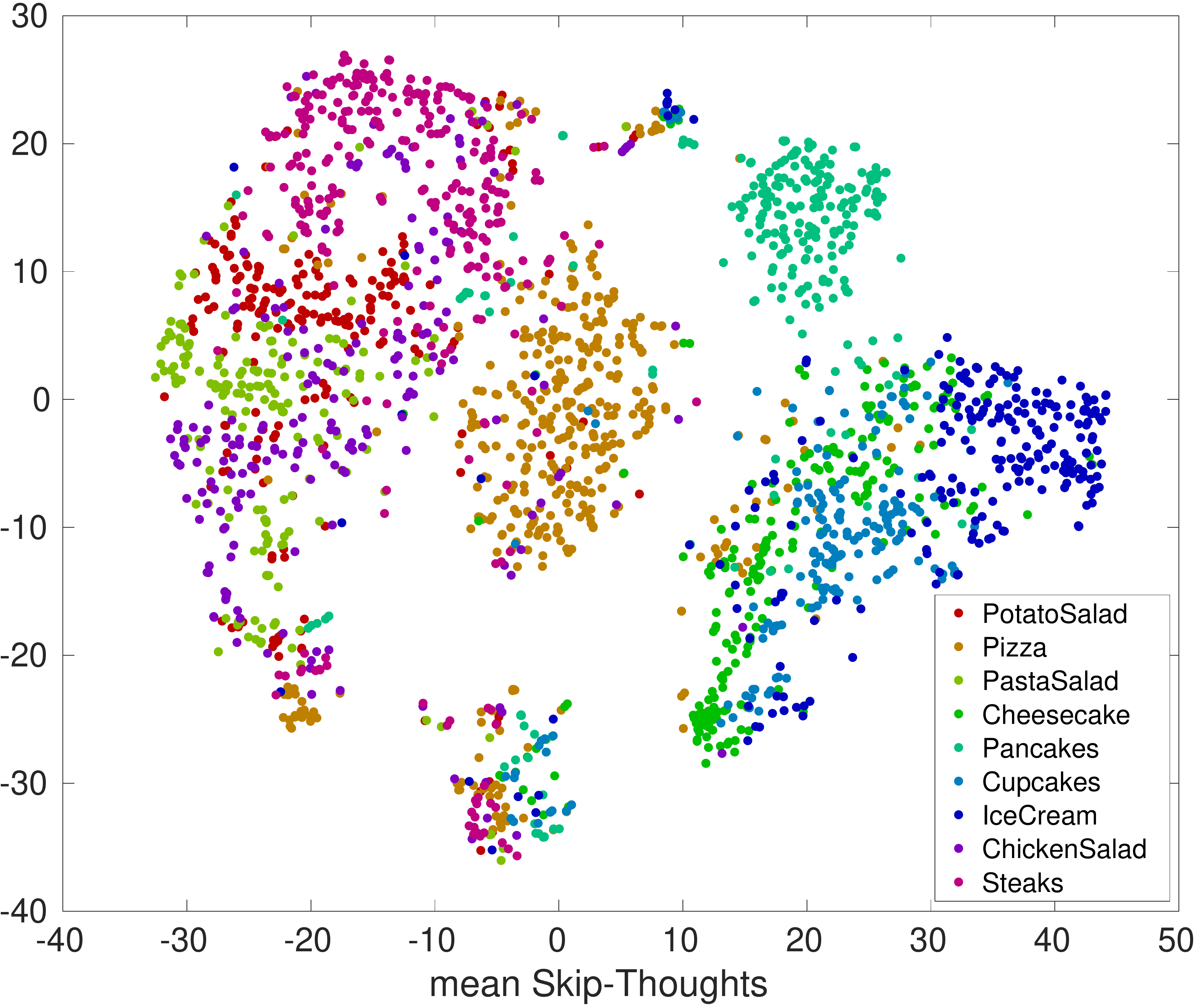} 
\vspace{-3.2mm}
\caption{Recipe encoding visualization with tSNE~\cite{Maaten2008VisualizingDU} over a set of recipes from the 9 most common categories in the Recipe1M test set.} 
\label{fig:tsne_9} 
\vspace{-4mm}
\end{figure*} 
\subsection{ Video Predictions}
\textbf{Window selection:} 
We test two video segmentation settings for inference: one according to ground truth (``Ours Visual (GT)'') and one based on fixed windows (``Ours Visual''). For ``Ours Visual'', we first partition the video until the last observation into fixed sized windows and sequentially feed these into our recipe RNN. Overall, our method is relatively robust to window size. We report results for different window sizes for YouCookII in Table~\ref{fig:Youcook2_window} and for Tasty in Table~\ref{fig:Tasty_window}.

\begin{table}[ht]
 \footnotesize
 \bgroup
 \def\arraystretch{1.25}
 \setlength\tabcolsep{0.255em}
 \setstretch{0.6}
 \begin{tabularx}{0.99\columnwidth}{|l|c|c|c|c|c|}
 \hline 
 Method & ING & verbs & BLEU1 & BLEU4 & METEOR \\ \hline
 Ours Visual (window 70) &\cellcolor[HTML]{FFEDF7}12.40 &\cellcolor[HTML]{F0FBF0}13.26 &\cellcolor[HTML]{FFF5EC}14.73 &\cellcolor[HTML]{FFECEC}0.93 &\cellcolor[HTML]{ECECFD}8.31 \\ \hline
 Ours Visual (window 90) &\cellcolor[HTML]{FFDCEF}13.15 &\cellcolor[HTML]{E8F9E8}13.99 &\cellcolor[HTML]{FFE8D0}15.97 &\cellcolor[HTML]{FFADAD}1.06 &\cellcolor[HTML]{C1C1F9}9.24 \\ \hline
 Ours Visual (window 110) &\cellcolor[HTML]{FFC0E2}14.06 &\cellcolor[HTML]{BDEFBD}15.58 &\cellcolor[HTML]{FFD7AF}16.84 &\cellcolor[HTML]{FF6262}1.18 &\cellcolor[HTML]{9696F5}10.05 \\ \hline
 Ours Visual (window 130) &\cellcolor[HTML]{FFA1D4}14.97 &\cellcolor[HTML]{C3F0C3}15.40 &\cellcolor[HTML]{FFBC79}17.94 &\cellcolor[HTML]{FF9B9B}1.09 &\cellcolor[HTML]{8888F3}10.32 \\ \hline
 Ours Visual (window 150) &\cellcolor[HTML]{FF60B6}16.70 &\cellcolor[HTML]{65DA65}17.59 &\cellcolor[HTML]{FF9F40}18.94 &\cellcolor[HTML]{FFA7A7}1.07 &\cellcolor[HTML]{5454EE}11.25 \\ \hline
 \textbf{Ours Visual (window 170)} &\cellcolor[HTML]{FF62B7}16.66 &\cellcolor[HTML]{7EE07E}17.08 &\cellcolor[HTML]{FFC58B}17.59 &\cellcolor[HTML]{FF4040}1.23 &\cellcolor[HTML]{6262EF}11.00 \\ \hline
 Ours Visual (window 190) &\cellcolor[HTML]{FF4FAE}17.14 &\cellcolor[HTML]{70DC70}17.38 &\cellcolor[HTML]{FFCF9F}17.18 &\cellcolor[HTML]{FF9B9B}1.09 &\cellcolor[HTML]{4040EC}11.60 \\ \hline
 Ours Visual (window 210) &\cellcolor[HTML]{FF7CC3}15.99 &\cellcolor[HTML]{B1ECB1}15.90 &\cellcolor[HTML]{FFC993}17.43 &\cellcolor[HTML]{FF5B5B}1.19 &\cellcolor[HTML]{6A6AF0}10.85 \\ \hline
 Ours Visual (window 230) &\cellcolor[HTML]{FF92CD}15.40 &\cellcolor[HTML]{C0F0C0}15.48 &\cellcolor[HTML]{FFE4C9}16.19 &\cellcolor[HTML]{FFADAD}1.06 &\cellcolor[HTML]{8888F3}10.31 \\ \hline
 \end{tabularx}
 \egroup
\vspace{-2mm}
 \caption{Window size selection on the Tasty Videos dataset } 
 \label{fig:Tasty_window}
 \vspace{-3mm}
\end{table}

\begin{table}[ht]
 \footnotesize
 \bgroup
 \def\arraystretch{1.25}
 \setlength\tabcolsep{0.295em}
 \setstretch{0.6}
 \begin{tabularx}{0.99\columnwidth}{|l|c|c|c|c|c|}
 \hline 
 Method & ING & verbs & BLEU1 & BLEU4 & METEOR \\ \hline 
 Ours Visual (window 30) &\cellcolor[HTML]{FFEDF7}15.18 &\cellcolor[HTML]{F0FBF0}20.38 &\cellcolor[HTML]{FFF5EC}19.99 &\cellcolor[HTML]{FFECEC}0.60 &\cellcolor[HTML]{ECECFD}9.21 \\ \hline
 Ours Visual (window 50) &\cellcolor[HTML]{FFD8ED}15.86 &\cellcolor[HTML]{D7F5D7}22.60 &\cellcolor[HTML]{FFDCB9}21.29 &\cellcolor[HTML]{FF8787}1.10 &\cellcolor[HTML]{A1A1F6}10.02 \\ \hline
 \textbf{Ours Visual (window 70)} &\cellcolor[HTML]{FF8CCA}17.64 &\cellcolor[HTML]{9DE79D}25.11 &\cellcolor[HTML]{FFB060}22.55 &\cellcolor[HTML]{FF4040}1.38 &\cellcolor[HTML]{5858EE}10.71 \\ \hline
 Ours Visual (window 90) &\cellcolor[HTML]{FF74BF}18.13 &\cellcolor[HTML]{79DE79}26.31 &\cellcolor[HTML]{FFA245}22.87 &\cellcolor[HTML]{FF5050}1.32 &\cellcolor[HTML]{4040EC}10.93 \\ \hline
 Ours Visual (window 110) &\cellcolor[HTML]{FF4FAE}18.86 &\cellcolor[HTML]{65DA65}26.91 &\cellcolor[HTML]{FF9F40}22.93 &\cellcolor[HTML]{FF5050}1.32 &\cellcolor[HTML]{4B4BED}10.83 \\ \hline
 Ours Visual (window 130) &\cellcolor[HTML]{FF70BD}18.21 &\cellcolor[HTML]{81E081}26.05 &\cellcolor[HTML]{FFB163}22.51 &\cellcolor[HTML]{FF5A5A}1.28 &\cellcolor[HTML]{4B4BED}10.83 \\ \hline
 Ours Visual (window 210) &\cellcolor[HTML]{FF78C1}18.05 &\cellcolor[HTML]{76DE76}26.40 &\cellcolor[HTML]{FFCB97}21.83 &\cellcolor[HTML]{FF6E6E}1.20 &\cellcolor[HTML]{B4B4F7}9.83 \\ \hline
 \end{tabularx}
 \egroup
\vspace{-2mm}
 \caption{Window size selection on the YouCookII dataset~\cite{zhou2018towards}. } 
 \label{fig:Youcook2_window}
 \vspace{-3mm}
\end{table}

\textbf{YouCookII dataset cross validation:} 
To benchmark our model on the YouCookII dataset, we create a zero-shot setting using 4-fold cross validation. We create our splits based on distinct dishes. First set includes all videos from dish labels between 1 and 125, second set 126 and 222, third set 223 and 316 and fourth set 317 and 425.

\begin{figure}[ht]
\centering 
\includegraphics[width=0.985\columnwidth]{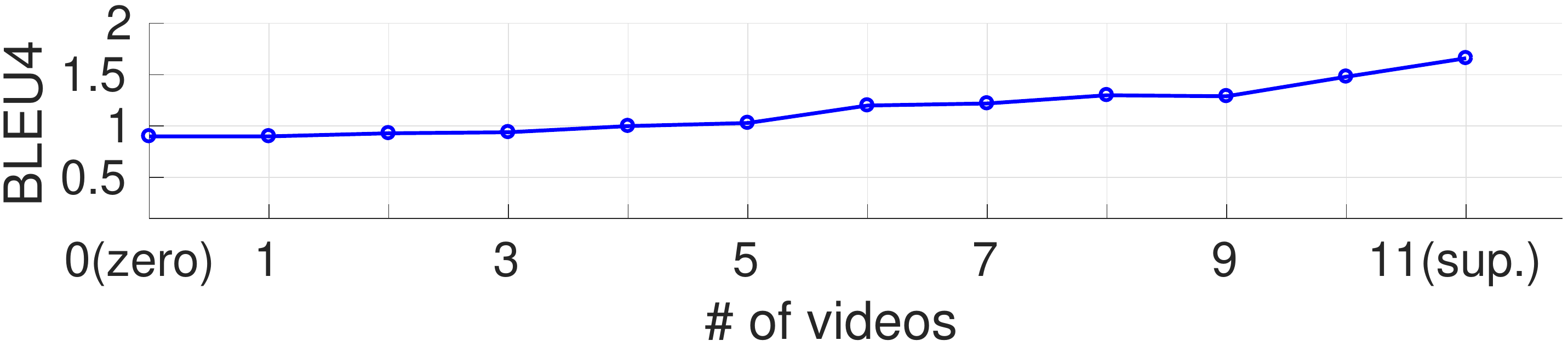} 
\caption{\footnotesize{BLEU4 increases with \# of training videos. 0/11 correspond to zero-shot/supervised settings; tested on YouCook2 set 1.}}
\label{fig:video_correction} 
\end{figure}

\textbf{YouCookII - more comparisons on supervised vs zero-shot performance:} 
YouCook2 averages 22 videos per dish. We used 11 for testing; for supervised-training we use the other 11 for training, but exclude them for a comparable zero-shot scenario.
Fig.~\ref{fig:video_correction} shows steady improvement when training with the dish-specific videos, indicating that the model is in fact learning and that more than 11 videos (current supervised setting) will further improve the supervised performance. 

\textbf{More results on the Tasty dataset:} 
We compare the prediction performance of visual model with GT segments vs. window segment in Figure~\ref{fig:tasty_all_res_video}. Compared to using ground truth segments, the fixed window segments do not have a significant decrease in performance.

 \vspace{-1mm}
\subsection{Ablation study} 
Since our method is modular, we conduct an ablation study to check the interchangeability of the sentence encoder on the Recipe1M dataset~\cite{salvador2017learning}. 
Instead of using our own sentence encoder, we represent the sentences using ST vectors trained on the Recipe1M dataset, as provided by the authors~\cite{salvador2017learning}. These vectors have been shown to perform well for their recipe retrieval. Our results, presented in Fig.~\ref{fig:im2recipe_percentage} show that our sentence encoder performs on par with ST encodings. Moreover, our encoder, model and decoder can all be trained jointly and do not require a separate pre-training of a sentence autoencoder. 
In both cases the recipe RNN and sentence decoder have been tested with the same parameter settings.
We also test how important the ingredients are as input for our method. We retrain our model without any ingredients using the same parameter settings.
Results are shown in Figure~\ref{fig:im2recipe_percentage}; we see that ingredient information is very important for our method, especially in predicting the initial steps. However, in subsequent steps when 25\%, 50\% of the recipe steps are seen, the model's performance starts to improve as it receives more information.

\subsection{Recipe Visualization} 
Our method can model recipes, as the output of the recipe RNN, especially after seeing all $N$ steps, serves as a feature vector representing the entire recipe. 
For validating these features we conduct a recipe visualization experiment.
We select recipes from the 9 most common recipe categories in the test set of the Recipe1M dataset~\cite{salvador2017learning} and encode them with our model by taking the final hidden output of the recipe RNN.
As comparison, similar to~\cite{salvador2017learning}, we take the mean of the ST vectors across the recipe steps. We visualize a two-dimensional representation computed using tSNE~\cite{Maaten2008VisualizingDU} of both recipe representations in Figure~\ref{fig:tsne_9}.
We find that with our method, the recipes are better separated according to category.

 \begin{figure*}[htb]
 \footnotesize
 \bgroup
 \def\arraystretch{1.25}
 \setlength\tabcolsep{0.35em}
 \setstretch{0.8}
 \begin{tabularx}{0.99\textwidth}{|c|X|X|c|c|c|c|c|}
 \hline
 & \textbf{ground truth} & \textbf{prediction} & \tiny{\textbf{BLEU1}} & \tiny{\textbf{BLEU4}} & \tiny{\textbf{METEOR}} & \tiny{\textbf{HUMAN1}}& \tiny{\textbf{HUMAN2}}\\\hline
 \textbf{ING} & salt, pepper, onion, egg, bacon, ketchup, ground beef, Worcestershire sauce, parmesan cheese, hamburger buns, shredded cheddar cheese & & & & & & \\\hline 
 \textbf{step1} & Preheat a grill for high heat. & In a large bowl, combine the first six ingredients. & \cellcolor{orange!11}11.0 & \cellcolor{red!0}0.0 & \cellcolor{blue!3}3.0 & 0.0 & \cellcolor{greenff!50}1.5 \\\hline
 \cellcolor{blue!5}\textbf{step2} & \cellcolor{blue!5}In a large bowl, mix together the Cheddar cheese, Parmesan cheese, onion, egg, ketchup, Worcestershire sauce, salt and pepper.& \cellcolor{blue!5}In a large bowl, mix together the ground beef, onion, Worcestershire sauce, egg, salt & \cellcolor{orange!66}66.0 & \cellcolor{red!38}38.0 &\cellcolor{blue!33}33.0 & \cellcolor{greenff!25}1.0 & \cellcolor{greenff!25}1.0 \\\hline
 \textbf{step3} & Crumble in the ground beef, and mix together by hand. & Shape into 4 patties. & \cellcolor{orange!0}0.0 & \cellcolor{red!0}0.0 & \cellcolor{blue!0} 0.0 & 0.0 & \cellcolor{greenff!25}1.0 \\\hline
 \cellcolor{lime!8}\textbf{step4} & \cellcolor{lime!8}Form into 6 patties, and wrap two slices of bacon around each one in both directions. & \cellcolor{lime!8}Form into 4 patties. & \cellcolor{orange!4}4.0 & \cellcolor{red!0}0.0 & \cellcolor{blue!10}10.0 & \cellcolor{greenff!10}0.5 & \cellcolor{greenff!50}1.5 \\\hline
 \textbf{step5} & Secure bacon with toothpicks. & Place on the grill, and cook for about 10 minutes, or until the bacon is cooked through & \cellcolor{orange!6}6.0 & \cellcolor{red!0}0.0 &\cellcolor{blue!9}9.0 & 0.0 & \cellcolor{greenff!25}1.0 \\\hline
 \cellcolor{blue!5}\textbf{step6} & \cellcolor{blue!5}Place patties on the grill, and cook for 5 minutes per side, or until well done. & \cellcolor{blue!5}Place on the grill, and cook for about 10 minutes, turning once. & \cellcolor{orange!48} 48.0 & \cellcolor{red!32}32.0 &\cellcolor{blue!22}22.0 & \cellcolor{greenff!50}1.5 & \cellcolor{greenff!50}1.5 \\\hline 
 \cellcolor{lime!8}\textbf{step7} & \cellcolor{lime!8}Remove toothpicks before serving on hamburger buns. & \cellcolor{lime!8}Serve on buns with lettuce, tomato, and ketchup. & \cellcolor{orange!25}25.0 & \cellcolor{red!0}0.0 &\cellcolor{blue!14}14.0 & \cellcolor{greenff!10}0.5 & \cellcolor{greenff!50}1.5 \\\hline
 \end{tabularx}
\egroup
\vspace{-1.5mm}
\caption{Predictions of our text-based method for ``Cheddar Bacon Wrapped Hamburgers'' along with the automated scores and human ratings. step4 prediction is half correct. step7 is a plausible prediction. } 
\vspace{-0.25cm}
\label{fig:example_preds_Cheddar}
\end{figure*} 

 \begin{figure*}[htb]
 \footnotesize
 \bgroup
 \def\arraystretch{1.25}
 \setlength\tabcolsep{0.25em}
 \setstretch{0.8}
 \begin{tabularx}{0.99\textwidth}{|c|X|X|c|c|c|c|c|}
 \hline
 & \textbf{ground truth} & \textbf{prediction} & \tiny{\textbf{BLEU1}} & \tiny{\textbf{BLEU4}} & \tiny{\textbf{METEOR}} & \tiny{\textbf{HUMAN1}}& \tiny{\textbf{HUMAN2}}\\\hline
 \textbf{ING} & milk, carrots, poultry seasoning, fresh ground black pepper, chicken bouillon cubes, celery ribs, boneless skinless chicken breasts & & & & & & \\\hline 
 \cellcolor{blue!5}\textbf{step1} & \cellcolor{blue!5}Place chicken in a slow cooker. & \cellcolor{blue!5}Place chicken in a large pot and cover with water. & \cellcolor{orange!40}40.0 & \cellcolor{red!26}26.0 & \cellcolor{blue!29}29.0 & \cellcolor{greenff!10}0.5 & \cellcolor{greenff!25}1.0 \\\hline
 \cellcolor{lime!8}\textbf{step2} & \cellcolor{lime!8}Heat broth in microwave ; dissolve bouillon in broth. & \cellcolor{lime!8}Add celery, carrots, and celery. & \cellcolor{orange!0}0.0 & \cellcolor{red!0}0.0 & \cellcolor{blue!0}0.0 & 0.0 & \cellcolor{greenff!70}2.0 \\\hline
 \textbf{step3} & Add next 4 ingredients to broth. & Pour over chicken. & \cellcolor{orange!0}0.0 & \cellcolor{red!0}0.0 & \cellcolor{blue!0}0.0 & 0.0 & \cellcolor{greenff!50}1.5 \\\hline
 \cellcolor{blue!5}\textbf{step4} & \cellcolor{blue!5}Pour over chicken. & \cellcolor{blue!5}Pour over chicken. & \cellcolor{orange!100}100 & \cellcolor{red!3}3.0 & \cellcolor{blue!80}100 & \cellcolor{greenff!70}2.0 & \cellcolor{greenff!70}2.0 \\\hline
 \cellcolor{lime!8}\textbf{step5} & \cellcolor{lime!8}Cover and cook on low for 6-8 hours, until chicken falls apart when poked with a fork. & \cellcolor{lime!8}Cover and cook on low for 8 hours. & \cellcolor{orange!28}28.0 & \cellcolor{red!23}23.0 & \cellcolor{blue!26}26.0 & \cellcolor{greenff!50}1.5 & \cellcolor{greenff!50}1.5 \\\hline
 \textbf{step6} & Combine buttermilk biscuit mix and milk, then drop spoonfuls over chicken to form dumplings. & Remove chicken from broth, cool and shred. & \cellcolor{orange!11}11.0 & \cellcolor{red!0}0.0 & \cellcolor{blue!5}5.0 & 0.0 & \cellcolor{greenff!25}1.0 \\\hline 
 \cellcolor{blue!5}\textbf{step7} & \cellcolor{blue!5}Cover and cook on high for 35 minutes or until dumplings are done. & \cellcolor{blue!5}Cover and cook on low for 8 to 10 hours or until chicken is tender. & \cellcolor{orange!47}47.0 & \cellcolor{red!20}20.0 & \cellcolor{blue!26}26.0 & \cellcolor{greenff!10}0.5 & \cellcolor{greenff!10}0.5 \\\hline
 \end{tabularx}
\egroup
\vspace{-1.5mm}
\caption{Predictions of our text-based method for ``Slow Cooker Chicken and Dumplings'' along with the automated scores and human ratings. step2 prediction is a plausible future step. step5 is correct. } 
\vspace{-0.25cm}
\label{fig:example_preds_Slow}
\end{figure*} 

 \begin{figure*}[htb]
 \footnotesize
 \bgroup
 \def\arraystretch{1.25}
 \setlength\tabcolsep{0.25em}
 \setstretch{0.8}
 \begin{tabularx}{0.99\textwidth}{|c|X|X|c|c|c|c|c|}
 \hline
 & \textbf{ground truth} & \textbf{prediction} & \tiny{\textbf{BLEU1}} & \tiny{\textbf{BLEU4}} & \tiny{\textbf{METEOR}} & \tiny{\textbf{HUMAN1}}& \tiny{\textbf{HUMAN2}}\\\hline
 \textbf{ING} & pineapple, strawberries, oranges, flaked coconut, fresh mint leaves, vanilla yogurt, kiwi fruits & & & & & & \\\hline 
 \cellcolor{lime!8}\textbf{step1} & \cellcolor{lime!8}Quarter pineapple lengthwise ; remove core. & \cellcolor{lime!8}In a large bowl, combine the strawberries, kiwi, oranges, and pineapple. & \cellcolor{orange!9}9.0 & \cellcolor{red!0}0.0 & \cellcolor{blue!7}7.0 & 0.0 & \cellcolor{greenff!25}1.0 \\\hline
 \cellcolor{blue!5}\textbf{step2} & \cellcolor{blue!5}Cut crosswise into small chunks. & \cellcolor{blue!5}Cut into 1/2-inch cubes. & \cellcolor{orange!39}39.0 & \cellcolor{red!0}0.0 & \cellcolor{blue!12}12.0 & \cellcolor{greenff!25}1.0 & \cellcolor{greenff!25}1.0 \\\hline
 \cellcolor{lime!8}\textbf{step3} & \cellcolor{lime!8}Place in large serving bowl. & \cellcolor{lime!8}In a large bowl, combine fruit, kiwi, and pineapple. & \cellcolor{orange!33}33.0 & \cellcolor{red!0}0.0 &\cellcolor{blue!20}20.0 & 0.0 & \cellcolor{greenff!25}1.0 \\\hline
 \textbf{step4} & Add clementine segments, strawberries, kiwifruit, and coconut ; gently toss. & Add fruit and nuts. & \cellcolor{orange!14}14.0 & \cellcolor{red!0}0.0 & \cellcolor{blue!7}7.0 & \cellcolor{greenff!10}0.5 & \cellcolor{greenff!10}0.5 \\\hline
 \textbf{step5} & Spoon into dessert glasses. & Serve immediately or store in refrigerator up to 3 days. &\cellcolor{orange!0} 0.0 & \cellcolor{red!0}0.0 &\cellcolor{blue!0}0.0 & 0.0 & \cellcolor{greenff!10}0.5 \\\hline
 \cellcolor{lime!8}\textbf{step6} & \cellcolor{lime!8}Top with a dollop of vanilla yogurt or sweetened sour cream. & \cellcolor{lime!8}Garnish with orange slices and mint. &\cellcolor{orange!7} 7.0 & \cellcolor{red!0}0.0 &\cellcolor{blue!2}2.0 & 0.0 & \cellcolor{greenff!25}1.0 \\\hline 
 \cellcolor{blue!5}\textbf{step7} & \cellcolor{blue!5}Garnish with mint sprigs if desired. & \cellcolor{blue!5}Garnish with orange slices and mint. &\cellcolor{orange!50} 50.0 & \cellcolor{red!0}0.0 &\cellcolor{blue!22}22.0 & \cellcolor{greenff!25}1.0 & \cellcolor{greenff!25}1.0 \\\hline
 \end{tabularx}
\egroup
\vspace{-1.5mm}
\caption{Predictions of our text-based method for ``Ambrosia Fruit Salad'' along with the automated scores and human ratings. step1, step3 and step6 are plausible future step predictions.} 
\vspace{-0.25cm}
\label{fig:example_preds_Ambrosia}
\end{figure*} 

 \begin{figure*}[htb]
 \footnotesize
 \bgroup
 \def\arraystretch{1.25}
 \setlength\tabcolsep{0.25em}
 \setstretch{0.8}
 \begin{tabularx}{0.99\textwidth}{|c|X|X|c|c|c|c|c|}
 \hline
 & \textbf{ground truth} & \textbf{prediction} & \tiny{\textbf{BLEU1}} & \tiny{\textbf{BLEU4}} & \tiny{\textbf{METEOR}} & \tiny{\textbf{HUMAN1}}& \tiny{\textbf{HUMAN2}}\\\hline
 \textbf{ING} & butter, cheese, eggs, salt and pepper & & & & & & \\\hline 
 \textbf{step1} & Whisk the eggs till light and fluffy. & Preheat oven to 350 degrees. & \cellcolor{orange!0}0.0 & \cellcolor{red!0}0.0 & \cellcolor{blue!0}0.0 & 0.0 & 0.0 \\\hline
 \cellcolor{blue!5}\textbf{step2} & \cellcolor{blue!5}Add cheese, salt and pepper. & \cellcolor{blue!5}Add the cheese and season with salt and pepper. & \cellcolor{orange!56}56.0 & \cellcolor{red!0}0.0 & \cellcolor{blue!43}43.0 & \cellcolor{greenff!70}2.0 & \cellcolor{greenff!70} 2.0 \\\hline
 \textbf{step3} & Mix well. & Melt butter in a pan. & \cellcolor{orange!0}0.0 & \cellcolor{red!0}0.0 & \cellcolor{blue!0}0.0 & 0.0 & \cellcolor{greenff!50}1.5 \\\hline
 \textbf{step4} & Heat a 7 inch non-stick skillet and add butter. & Melt butter in a pan. & \cellcolor{orange!18}18.0 & \cellcolor{red!0}0.0 & \cellcolor{blue!11}11.0 & \cellcolor{greenff!25} 1.0 & \cellcolor{greenff!25}1.0 \\\hline
 \cellcolor{lime!8}\textbf{step5} & \cellcolor{lime!8}As the butter begins to smoke, lower heat and pour the beaten egg mixture. & \cellcolor{lime!8}Pour in egg mixture. & \cellcolor{orange!6}6.0 & \cellcolor{red!0}0.0 & \cellcolor{blue!14}14.0 & \cellcolor{greenff!25}1.0 & \cellcolor{greenff!25}1.0 \\\hline
 \cellcolor{blue!5}\textbf{step6} & \cellcolor{blue!5}Cook covered for a minute. & \cellcolor{blue!5}Let cook until eggs are set. & \cellcolor{orange!17}17.0 & \cellcolor{red!0}0.0 & \cellcolor{blue!10}10.0 & \cellcolor{greenff!50}1.5 & \cellcolor{greenff!50}1.5 \\\hline 
 \cellcolor{lime!8}\textbf{step7} & \cellcolor{lime!8}Fold over and serve immediately. & \cellcolor{lime!8}Turn the heat down to low and cook for about 5 minutes. & \cellcolor{orange!8}8.0 & \cellcolor{red!0}0.0 &\cellcolor{blue!3}3.0 & 0.0 & \cellcolor{greenff!25}1.0 \\\hline
 \end{tabularx}
\egroup
\vspace{-1.5mm}
\caption{Predictions of our text-based method for ``Cheese Omelette'' along with the automated scores and human ratings. step5 is a correct prediction. step7 is a plausible prediction.} 
\vspace{-0.25cm}
\label{fig:example_preds_Cheese}
\end{figure*} 

 \begin{figure*}[htb]
 \footnotesize
 \bgroup
 \def\arraystretch{1.25}
 \setlength\tabcolsep{0.25em}
 \setstretch{0.8}
 \begin{tabularx}{0.99\textwidth}{|c|X|X|c|c|c|c|c|}
 \hline
 & \textbf{ground truth} & \textbf{prediction} & \tiny{\textbf{BLEU1}} & \tiny{\textbf{BLEU4}} & \tiny{\textbf{METEOR}} & \tiny{\textbf{HUMAN1}}& \tiny{\textbf{HUMAN2}}\\\hline
 \textbf{ING} & sugar, celery, black olives, chopped tomatoes, green olives & & & & & & \\\hline 
 \textbf{step1} & Boil your noodles as directed on box. & Mix all ingredients together. & \cellcolor{orange!0}0.0 & \cellcolor{red!0}0.0 & \cellcolor{blue!0}0.0 & 0.0 & \cellcolor{greenff!10} 0.5 \\\hline
 \cellcolor{lime!8}\textbf{step2} & \cellcolor{lime!8}Do not overcook! & \cellcolor{lime!8}Drain and rinse with cold water. & \cellcolor{orange!0}0.0 & \cellcolor{red!0}0.0 & \cellcolor{blue!0}0.0 & 0.0 & \cellcolor{greenff!70}2.0 \\\hline
 \cellcolor{blue!5}\textbf{step3} & \cellcolor{blue!5}Drain. & \cellcolor{blue!5}Drain and rinse with cold water. & \cellcolor{orange!17}17.0 & \cellcolor{red!0}0.0 & \cellcolor{blue!26}26.0 & \cellcolor{greenff!50}1.5 & \cellcolor{greenff!50}1.5 \\\hline
 \cellcolor{lime!8}\textbf{step4} & \cellcolor{lime!8}Add veggies to pasta in large bowl. & \cellcolor{lime!8}In a large bowl, mix together the noodles, celery, tomatoes, olives, and sugar. &\cellcolor{orange!23}23.0 & \cellcolor{red!0}0.0 & \cellcolor{blue!16}16.0 & \cellcolor{greenff!50}1.5 & \cellcolor{greenff!50}1.5 \\\hline
 \textbf{step5} & Add Italian dressing and Splenda or sugar and ground pepper. & Add all other ingredients. & \cellcolor{orange!6}6.0 & \cellcolor{red!0}0.0 & \cellcolor{blue!6}6.0 & \cellcolor{greenff!50}1.5 & \cellcolor{greenff!50}1.5 \\\hline
 \cellcolor{blue!5}\textbf{step6} & \cellcolor{blue!5}Mix well, chill and enjoy! & \cellcolor{blue!5}Mix well. &\cellcolor{orange!22}22.0 & \cellcolor{red!0}0.0 & \cellcolor{blue!24}24.0 & \cellcolor{greenff!50}1.5 & \cellcolor{greenff!50}1.5 \\\hline 
 \cellcolor{lime!8}\textbf{step7} & \cellcolor{lime!8}Add more dressing the next day as needed, if put in the fridge overnight! & \cellcolor{lime!8}I like to add a little bit of olive oil to the salad and I add a little more & \cellcolor{orange!16}16.0 & \cellcolor{red!0}0.0 &\cellcolor{blue!6}6.0 & \cellcolor{greenff!10}0.5 & \cellcolor{greenff!25}1.0 \\\hline
 \end{tabularx}
\egroup
\vspace{-1.5mm}
\caption{Predictions of our text-based method for ``My Mom's Summer Pasta Salad'' along with the automated scores and human ratings. step2 is a plausible future step. step4 is correct as veggies refer to celery tomatoes and olives. step7 is a plausible suggestion even though it does not match the GT step. } 
\vspace{-0.25cm}
\label{fig:example_preds_Summer}
\end{figure*} 

 \begin{figure*}[htb]
 \footnotesize
 \bgroup
 \def\arraystretch{1.25}
 \setlength\tabcolsep{0.25em}
 \setstretch{0.8}
 \begin{tabularx}{0.99\textwidth}{|c|X|X|c|c|c|c|c|}
 \hline
 & \textbf{ground truth} & \textbf{prediction} & \tiny{\textbf{BLEU1}} & \tiny{\textbf{BLEU4}} & \tiny{\textbf{METEOR}} & \tiny{\textbf{HUMAN1}}& \tiny{\textbf{HUMAN2}}\\\hline
 \textbf{ING} &salt and pepper, parmesan cheese, garlic cloves, extra virgin olive oil, diced tomatoes, fusilli, pesto sauce, fresh basil leaves & & & & & & \\\hline 
 \cellcolor{blue!5}\textbf{step1} & \cellcolor{blue!5}Heat oil in a medium skillet over medium heat. & \cellcolor{blue!5}Heat the oil in a large skillet over medium heat. & \cellcolor{orange!80} 80.0 & \cellcolor{red!39}39.0 & \cellcolor{blue!43}43.0 & \cellcolor{greenff!70}2.0 & \cellcolor{greenff!70}2.0 \\\hline
 \textbf{step2} & Saute garlic for 2 minutes, then add tomatoes and simmer for 10 minutes. & Add garlic and cook until fragrant, about 30 seconds. & \cellcolor{orange!21} 21.0 & \cellcolor{red!0}0.0 &\cellcolor{blue!9} 9.0 & \cellcolor{greenff!10} 0.5 & \cellcolor{greenff!50}1.5 \\\hline
 \cellcolor{lime!8}\textbf{step3} & \cellcolor{lime!8}Meanwhile, bring a large pot of lightly salted water to a boil. & \cellcolor{lime!8}Add pesto and cook for 1 minute. & \cellcolor{orange!0}0.0 & \cellcolor{red!0}0.0 &\cellcolor{blue!4}4.0 & 0.0 & \cellcolor{greenff!50}1.5 \\\hline
 \cellcolor{blue!5}\textbf{step4} & \cellcolor{blue!5}Add pasta and cook for 8 to 10 minutes or until al dente ; drain and return to pot. & \cellcolor{blue!5}Add pasta and cook for 8 to 10 minutes or until al dente ; drain. & \cellcolor{orange!75}75.0 & \cellcolor{red!75}75.0 & \cellcolor{blue!54}54.0 & \cellcolor{greenff!70} 2.0 & \cellcolor{greenff!70}2.0 \\\hline
 \cellcolor{lime!8}\textbf{step5} & \cellcolor{lime!8} Season tomato sauce with salt and pepper ; stir in pesto. & \cellcolor{lime!8}Add pesto and basil to pasta and toss to combine. & \cellcolor{orange!20}20.0 & \cellcolor{red!0}0.0 & \cellcolor{blue!7}7.0& \cellcolor{greenff!10}0.5 & \cellcolor{greenff!50}1.5 \\\hline
 \cellcolor{blue!5}\textbf{step6} & \cellcolor{blue!5}Pour sauce over hot cooked pasta and mix well. & \cellcolor{blue!5}Pour sauce over pasta and toss to coat. & \cellcolor{orange!55}55.0 & \cellcolor{red!0}0.0 &\cellcolor{blue!25}25.0 & \cellcolor{greenff!25}1.0 & \cellcolor{greenff!25}1.0 \\\hline 
 \textbf{step7} & Sprinkle with cheese, garnish with basil leaves and serve immediately. & Serve immediately. & \cellcolor{orange!2}2.0 & \cellcolor{red!0}0.0 &\cellcolor{blue!14}14.0 & \cellcolor{greenff!25}1.0 & \cellcolor{greenff!25}1.0 \\\hline
 \end{tabularx}
\egroup
\vspace{-1.5mm}
\caption{Predictions of our text-based method for ``Easy Fusilli with Tomato Pesto Sauce'' along with the automated scores and human ratings. step3 is a plausible future prediction. step5 is half correct.} 
\vspace{-0.25cm}
\label{fig:example_preds_Fusilli}
\end{figure*} 
 
 \begin{figure*}[htb]
 \footnotesize
 \bgroup
 \def\arraystretch{1.25}
 \setlength\tabcolsep{0.25em}
 \setstretch{0.8}
 \begin{tabularx}{0.99\textwidth}{|c|X|X|c|c|c|c|c|}
 \hline
 & \textbf{ground truth} & \textbf{prediction} & \tiny{\textbf{BLEU1}} & \tiny{\textbf{BLEU4}} & \tiny{\textbf{METEOR}} & \tiny{\textbf{HUMAN1}}& \tiny{\textbf{HUMAN2}}\\\hline
 \textbf{ING} & salt, butter, cheese, cream cheese, scallions, cauliflower, black pepper, bacon bits, whole milk& & & & & & \\\hline 
 \cellcolor{blue!5}\textbf{step1} & \cellcolor{blue!5}Preheat oven to 350 & \cellcolor{blue!5}Preheat oven to 350 degrees F. & \cellcolor{orange!67} 67.0 &\cellcolor{red!51} 51.0 & \cellcolor{blue!50}50.0 & \cellcolor{greenff!70} 2.0 & \cellcolor{greenff!70}2.0 \\\hline
 \textbf{step2} & Chop cauliflower and boil in salted water for about 3-4 mins until a little tender & Place cauliflower in a large pot and cover with water. & \cellcolor{orange!30} 30.0 & \cellcolor{red!0}0.0 & \cellcolor{blue!10}10.0 & \cellcolor{greenff!25}1.0 & \cellcolor{greenff!25}1.0 \\\hline
 \textbf{step3} & Strain cauliflower and mash with a potato masher. & Add the butter, salt, and pepper, and stir to combine. &\cellcolor{orange!10}10.0 & \cellcolor{red!0}0.0 &\cellcolor{blue!2}2.0 & 0.0 & \cellcolor{greenff!50}1.5 \\\hline
 \textbf{step4} & Place in bowl and mix all ingredients & Melt butter in a large skillet over medium heat. & \cellcolor{orange!11}11.0 & \cellcolor{red!0}0.0 & \cellcolor{blue!3}3.0 & 0.0 & \cellcolor{greenff!25}1.0 \\\hline
 \cellcolor{lime!8}\textbf{step5} & \cellcolor{lime!8}Add ingredients to a buttered 8x8 dish and top with extra cheese and bacon. & \cellcolor{lime!8} Pour mixture into a baking dish and bake for 20 minutes &\cellcolor{orange!21} 21.0 & \cellcolor{red!0}0.0 & \cellcolor{blue!7}7.0 & \cellcolor{greenff!10}0.5 & \cellcolor{greenff!50}1.5 \\\hline
 \cellcolor{blue!5}\textbf{step6} & \cellcolor{blue!5}Bake for 30 mins until nice and bubbly. & \cellcolor{blue!5}Bake for 30 minutes. & \cellcolor{orange!27}27.0 & \cellcolor{red!0}0.0 &\cellcolor{blue!28}28.0 & \cellcolor{greenff!70} 2.0 & \cellcolor{greenff!70} 2.0 \\\hline 
 \textbf{step7} & Enjoy! & Serve with a side of buttered toast. & \cellcolor{orange!0}0.0 & \cellcolor{red!0}0.0 &\cellcolor{blue!0}0.0 & \cellcolor{greenff!10}0.5 & \cellcolor{greenff!50}1.5 \\\hline
 \end{tabularx}
 \egroup
\vspace{-1.5mm}
\caption{Predictions of our text-based method for ``Baked Loaded Cauliflower'' along with the automated scores and human ratings. step5 is half correct.} 
\vspace{-0.25cm}
\label{fig:example_preds_Baked}
\end{figure*} 
 
 \begin{figure*}[htb]
 \footnotesize
 \bgroup
 \def\arraystretch{1.25}
 \setlength\tabcolsep{0.25em}
 \setstretch{0.8}
 \begin{tabularx}{0.99\textwidth}{|c|X|X|c|c|c|c|c|}
 \hline
 & \textbf{ground truth} & \textbf{prediction} & \tiny{\textbf{BLEU1}} & \tiny{\textbf{BLEU4}} & \tiny{\textbf{METEOR}} & \tiny{\textbf{HUMAN1}}& \tiny{\textbf{HUMAN2}}\\\hline
 \textbf{ING} & salt, butter, potatoes, sour cream, green onions, sharp cheddar cheese & & & & & & \\\hline 
 \cellcolor{lime!8}\textbf{step1} & \cellcolor{lime!8}Peel, quarter, and boil potatoes until barely tender. & \cellcolor{lime!8}Preheat oven to 350 degrees. & \cellcolor{orange!0}0.0 & \cellcolor{red!0}0.0 & \cellcolor{blue!0}0.0 & 0.0 & \cellcolor{greenff!70}2.0 \\\hline
 \textbf{step2} & Drain, cool and grate on large side of grater. & Melt butter in a large skillet over medium heat. & \cellcolor{orange!11}11.0 & \cellcolor{red!0}0.0 &\cellcolor{blue!6} 6.0 & 0.0 & \cellcolor{greenff!70}2.0 \\\hline
 \textbf{step3} & Place in a large bowl. & In a large skillet, melt butter over medium heat. & \cellcolor{orange!33}33.0 & \cellcolor{red!0}0.0 & \cellcolor{blue!21}21.0 & 0.0 & \cellcolor{greenff!70}2.0 \\\hline
 \cellcolor{blue!5}\textbf{step4} & \cellcolor{blue!5}Gently mix in all other ingredients, except butter. & \cellcolor{blue!5}Add remaining ingredients and mix well. & \cellcolor{orange!24}24.0 & \cellcolor{red!0}0.0 & \cellcolor{blue!17}17.0 & \cellcolor{greenff!25}1.0 & \cellcolor{greenff!25}1.0 \\\hline
 \textbf{step5} & Place mixture into a greased shallow 9x9 pan. & Spoon mixture into a lightly greased 9 x 13 inch baking dish. & \cellcolor{orange!33}33.0 & \cellcolor{red!0}0.0 & \cellcolor{blue!18}18.0 & \cellcolor{greenff!50}1.5 & \cellcolor{greenff!50}1.5 \\\hline
 \cellcolor{lime!8}\textbf{step6} & \cellcolor{lime!8}Top with melted butter. & \cellcolor{lime!8}Bake at 350 degrees for 30 minutes. & \cellcolor{orange!0}0.0 & \cellcolor{red!0}0.0 & 0.0 & \cellcolor{blue!0}0.0 & \cellcolor{greenff!50}1.5 \\\hline 
 \cellcolor{blue!5}\textbf{step7} & \cellcolor{blue!5}Bake at 400 degrees for 30 minutes or until golden. & \cellcolor{blue!5}Bake at 350 degrees for 1 hour. & \cellcolor{orange!37}37.0 & \cellcolor{red!0}0.0 &\cellcolor{blue!21}21.0 & \cellcolor{greenff!50}1.5 & \cellcolor{greenff!50}1.5 \\\hline
 \end{tabularx}
\egroup
\vspace{-1.5mm}
\caption{Predictions of our text-based method for ``Green Onion Potato Casserole'' along with the automated scores and human ratings. step1 is a plausible prediction as the oven will be used for baking. step6 is a plausible future step. } 
\vspace{-0.25cm}
\label{fig:example_preds_Green}
\end{figure*}

\begin{figure}[ht]
\centering 
\includegraphics[width=\columnwidth]{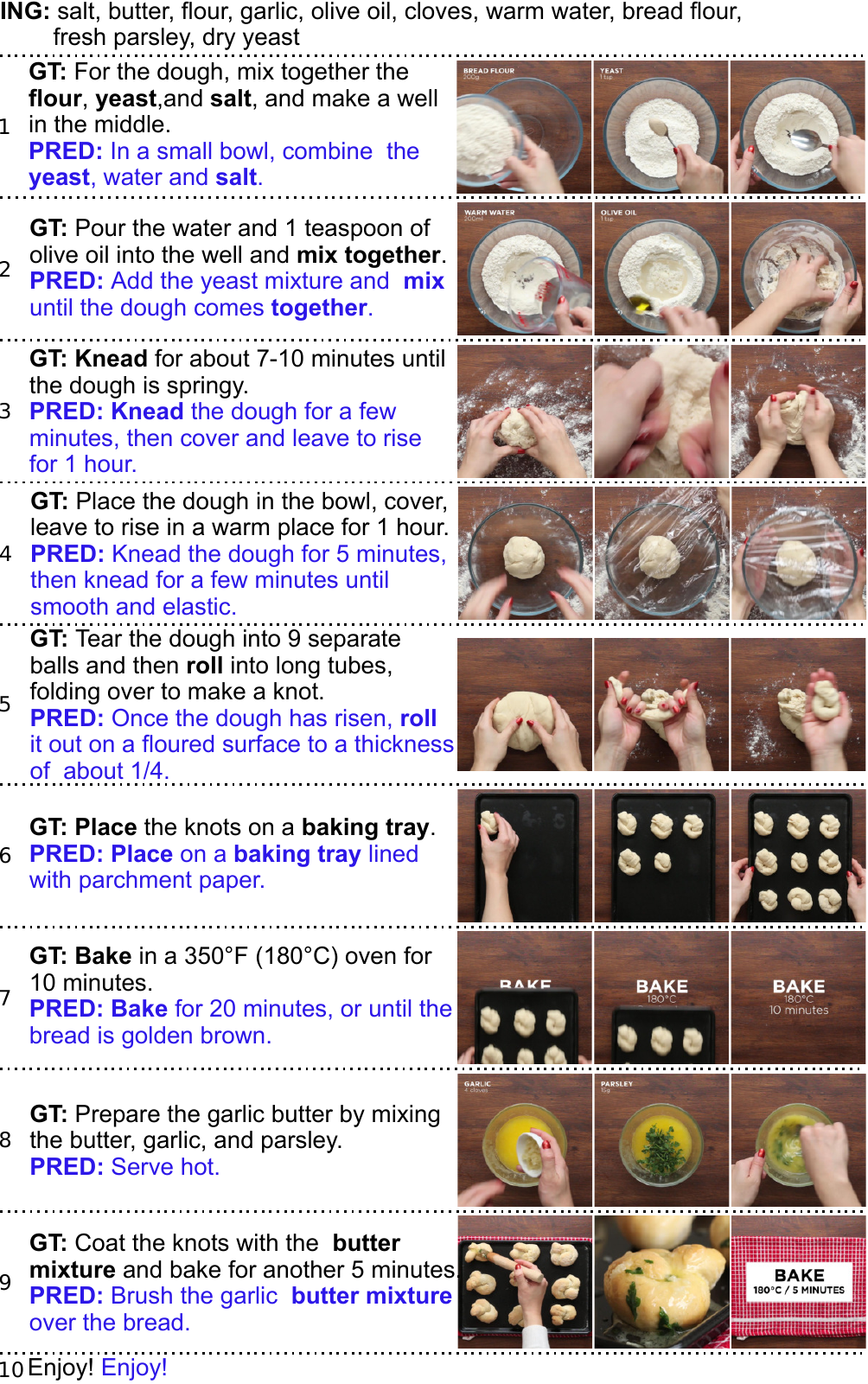}
\vspace{-1.8mm}
\caption{Next step prediction of our visual model for ``Garlic Knots''. The blue sentences are our model's predictions. After baking, in step7, our model predicts that the dish should be served, but after visually seeing the butter parsley mixture in step8, it correctly predicts that the knots should be brushed in step9. 
Note that our model predicts the next steps before seeing these segments! }
\vspace{-0.3cm}
\label{fig:videoexp_Garlic} 
\end{figure} 

\begin{figure}[ht]
\centering 
\includegraphics[width=\columnwidth]{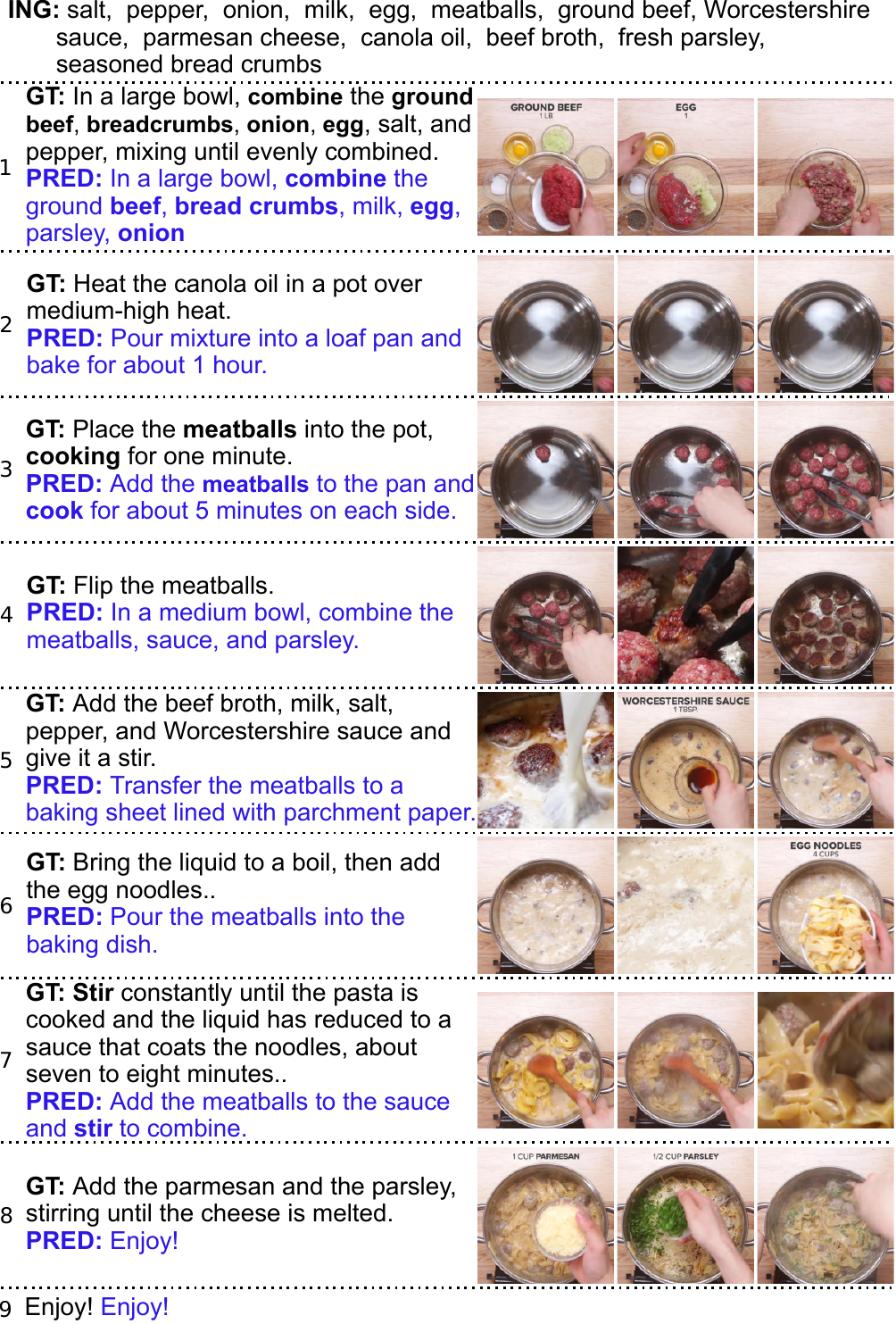}
\vspace{-1.8mm}
\caption{Next step prediction of our visual model for ``One-pot Swedish Meatball Pasta''. The blue sentences are our model's predictions. 
Our prediction for step3 matches the GT step4.
Our model's prediction for step4 is somehow plausible as a future step as the GT in step5 suggest mixing the sauce and meatballs. 
Note that our model predicts the next steps before seeing these segments!}
\vspace{-0.3cm}
\label{fig:videoexp_Swedish} 
\end{figure} 

\begin{figure}[ht]
\centering 
\includegraphics[width=\columnwidth]{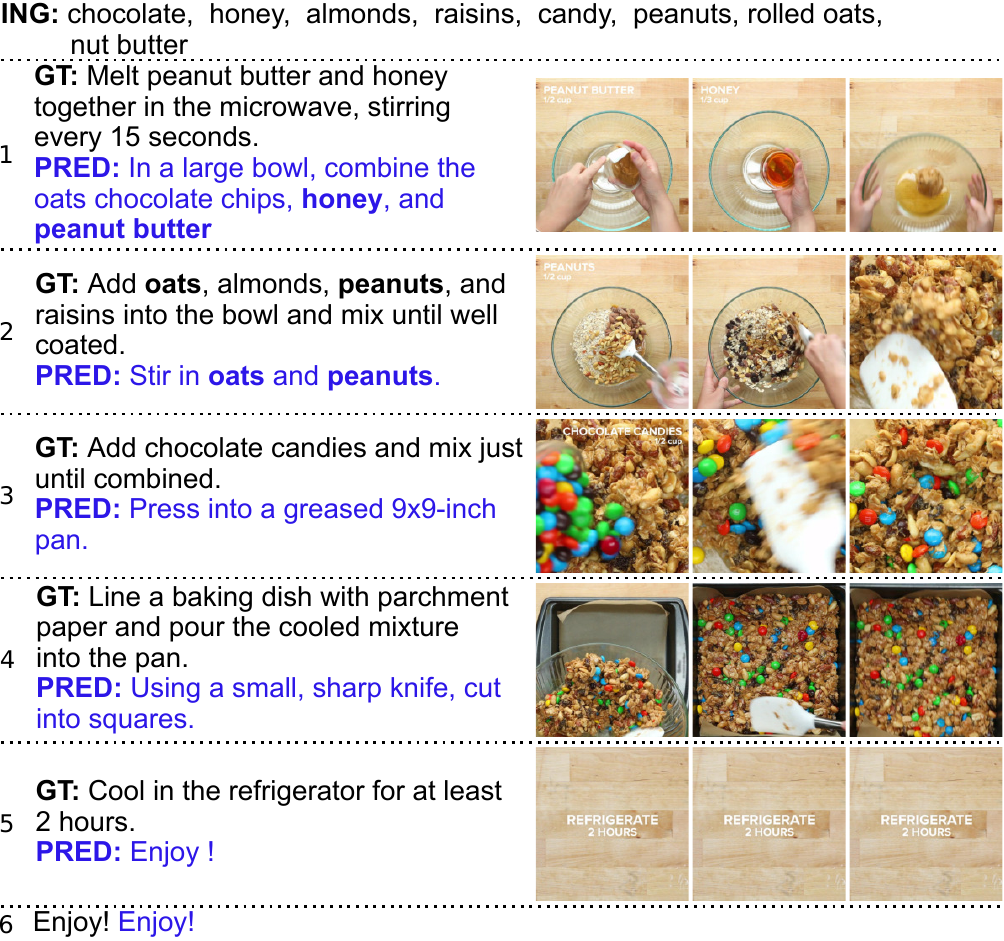}
\vspace{-1.8mm}
\caption{Next step prediction of our visual model for ``Trail Mix Bars''. The blue sentences are our model's predictions. Our model's prediction for step3 is a plausible future step prediction as the mixture is placed in a pan in step4. Note that our model predicts the next steps before seeing these segments!}
\vspace{-0.3cm}
\label{fig:videoexp_Trail} 
\end{figure} 

{\small
\bibliographystyle{ieee}
\bibliography{supplement}
}